\documentclass[sigconf]{acmart}

\usepackage{booktabs} 
\usepackage{epsfig}
\usepackage{graphicx}
\usepackage{amsmath}
\usepackage{amssymb}
\usepackage{multirow}

\setcopyright{acmcopyright}

\begin{document}
\title{Fully Point-wise Convolutional Neural Network for Modeling Statistical Regularities in Natural Images}

\author{Jing Zhang}
\affiliation{%
  \institution{Hangzhou Dianzi University}
  \city{Hangzhou}
  \country{China}}
 \email{jingzhang@hdu.edu.cn}

\author{Yang Cao}
\authornote{Corresponding author: forrest@ustc.edu.cn}
\affiliation{%
  \institution{University of Science and Technology of China}
  \city{Hefei}
  \country{China}
}

\author{Yang Wang}
\affiliation{%
 \institution{University of Science and Technology of China}
 \city{Hefei}
 \country{China}
}

\author{Chenglin Wen}
\affiliation{%
  \institution{Hangzhou Dianzi University}
  \city{Hangzhou}
  \country{China}
}

\author{Chang Wen Chen}
\affiliation{%
  \institution{University at Buffalo, The State University of New York}
  \city{Buffalo}
  \country{U.S.A.}
}

\renewcommand{\shortauthors}{J. Zhang et al.}

\begin{abstract}
Modeling statistical regularity plays an essential role in ill-posed image processing problems. Recently, deep learning based methods have been presented to implicitly learn statistical representation of pixel distributions in natural images and leverage it as a constraint to facilitate subsequent tasks, such as color constancy and image dehazing. However, the existing CNN architecture is prone to variability and diversity of pixel intensity within and between local regions, which may result in inaccurate statistical representation. To address this problem, this paper presents a novel fully point-wise CNN architecture for modeling statistical regularities in natural images. Specifically, we propose to randomly shuffle the pixels in the origin images and leverage the shuffled image as input to make CNN more concerned with the statistical properties. Moreover, since the pixels in the shuffled image are independent identically distributed, we can replace all the large convolution kernels in CNN with point-wise ($1*1$) convolution kernels while maintaining the representation ability. Experimental results on two applications: color constancy and image dehazing, demonstrate the superiority of our proposed network over the existing architectures, i.e., using 1/10$\sim$1/100 network parameters and computational cost while achieving comparable performance.
\end{abstract}

\copyrightyear{2018}
\acmYear{2018}
\setcopyright{acmcopyright}
\acmConference[MM '18]{2018 ACM Multimedia Conference}{October 22--26, 2018}{Seoul, Republic of Korea}
\acmBooktitle{2018 ACM Multimedia Conference (MM '18), October 22--26, 2018, Seoul, Republic of Korea}
\acmPrice{15.00}
\acmDOI{10.1145/3240508.3240653}
\acmISBN{978-1-4503-5665-7/18/10}

%
%



\maketitle


\section{Introduction}
Modeling statistical regularities is essential for natural image processing because of its impact on solving the ill-posed problem. Due to the complex, diverse and high-dimensional distributions of the pixels, it is still a challenging task to discover and model the statistical regularities in natural images.

\begin{figure*}[t]
\begin{center}
\includegraphics[width=1\linewidth]{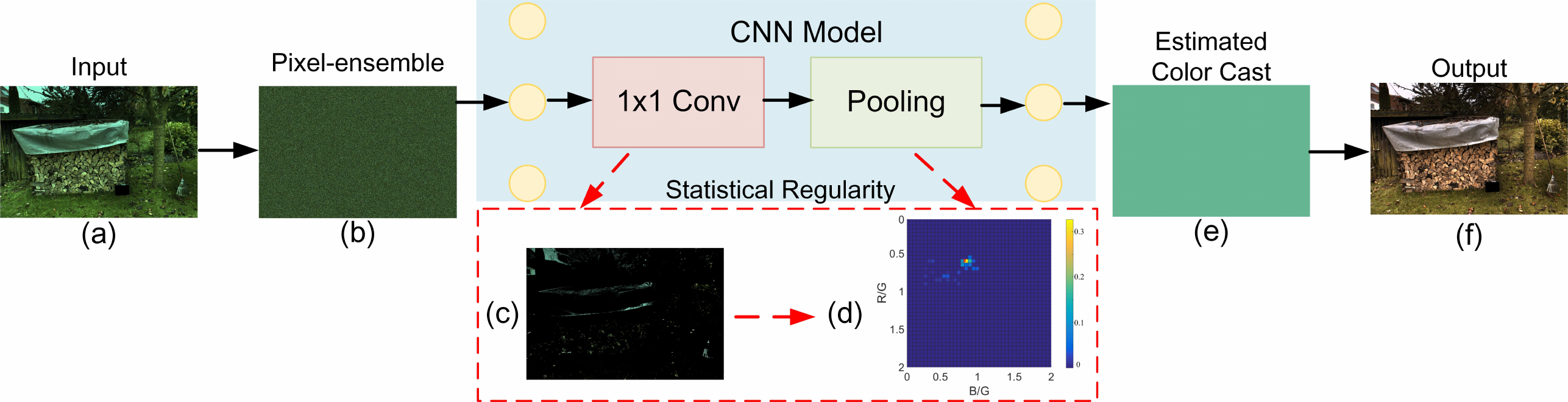}
\end{center}
   \caption{An exemplar illustration of the proposed method for color constancy. (a) An illuminated image. (b)Network input: a pixel-ensemble generated by shuffling (a). (c) Re-projection of the first pooling layer response map onto (a). (d) Bird view of the weighted histogram of (c). The red cross denotes the ground truth color cast (R/G:0.59, B/G:0.83). Warm color represents high frequency. (e) Color cast prediction. (f) The restored intrinsic image using predicted color cast. Best viewed on screen.}
\label{fig:fig1-illustration}
\end{figure*}

One feasible solution is to assume the statistical regularities, which utilizes the prior knowledge about the types of distributions that exist, and thus to design specialized algorithms for the subsequent tasks. For example, many color constancy algorithms work by assuming some regularities in the colors of natural objects viewed under canonical illumination, e.g., gray world \cite{buchsbaum1980spatial}, gray edge \cite{van2007edge}, and shades of gray \cite{finlayson2004shades}. Besides, by assuming that the surface shading and scene transmission are locally uncorrelated, most of single image dehazing methods are proposed based on various image priors, e.g., color attenuation \cite{zhu2014single}, dark channel \cite{CVPR2009_He, he2011single}, haze line \cite{berman2016non} and maximum reflectance prior \cite{zhang2017fast}.

Another practicable approach is to learn the statistical regularities, which formulates regression models of how the pixels are distributed, and adapts the model parameters to fit the input images. Thus, the adapted model parameters indeed reveal the statistics of the pixels, while the internal representations of models reflect the individual pixel patterns. This approach makes minimal assumptions on the pixel distributions and results in more general representations \cite{karklin2005hierarchical}.

Recently, deep learning has made much success in natural image processing problems like image denoising \cite{burger2012image, xie2012image}, super-resolution \cite{dong2016image} and the most relevant ones with statistical regularities, e.g. color constancy \cite{shi2016deep} and image dehazing \cite{cai2016dehazenet}. The hierarchical neural network representations of deep learning can make the regression task much simpler, and learn the sub-models of regularities and their corresponding active regions jointly. Building convolutional neural networks (CNNs) with better representation ability become popular in the development of major image processing methods.

However, the existing CNN architectures are more concerned with the local structural features caused by the variability and diversity of pixel intensity within and between local regions. These structural features might provide discriminative cues for visual recognition. However, the inconsistency between the structural features of local patch will introduce interferences into formulating a unified representation of statistical regularities. Moreover, to achieve better modeling capacity, the existing deep learning methods tend to pursuit a deeper and larger CNN. For example, the powerful CNNs for image processing tasks usually have dozens of layers and hundreds of channels \cite{shi2016deep, hu2017fc, zhang2017beyond}, thus resulting in millions of network parameters. This leads to a high computational cost, which limits its wide range of applications.

In this paper, we present a novel fully point-wise CNN architecture for modeling statistical regularities in natural images. We find that, shuffling the pixels in an image patch does not change their statistics while destroying the spatial structures. Inspired by this observation, we propose to use the shuffled image as input of CNN in order to facilitate the modeling of statistical properties. We show that, since the pixels in the shuffled image are independent identically distributed (i.i.d.), the statistical regularities of the original input image are, i) well preserved in the obtained pixel ensembles, and ii) able to be represented by using $1*1$ (point-wise) convolution kernels instead of $k*k$ convolution kernels. Accordingly, we propose a novel CNN architecture consisting of fully point-wise convolution units. This network can greatly reduce the network complexity while maintaining representation capability. Compared with the existing CNN architecture for image processing applications, our proposed architecture is lightweight, compact and resisting to overfitting.

A typical example of our proposed network performing on color constancy application is shown in Fig. 1. As can be seen, statistics of color cast are well preserved in the shuffled image, and more easily to be represented by our proposed fully point-wise CNN. As revealed in Fig.~\ref{fig:fig1-illustration}(c) and (d), the statistical regularity is implicitly modeled by sampling important pixels according to the neuron activations. Then the regularity is leveraged to achieve an efficient estimate of the color cast, determining a global color constancy result. Various experimental results further demonstrate the superiority of our proposed network over the existing architectures.

The main contribution is that, to the best of our knowledge, we are the first to propose a \emph{fully} point-wise CNN architecture for modeling statistical regularities in natural images. We present a pixel shuffling strategy to make CNN more concerned with the statistical properties in the input images. Our proposed CNN architecture is lightweight, compact and resisting to overfitting. Generally, it only needs 1/10$\sim$1/100 parameters and computational cost over the state-of-the-art networks while maintaining comparable accuracy.

\section{Related Work}
\subsection{Modeling statistical regularities}
Modeling statistical regularities is an important topic in natural image processing. A comprehensive review about literatures in this topic is beyond the scope of this paper. Here we choose the typical image enhancement applications that are most relevant with statistical regularities, i.e., color constancy and image dehazing, and present a brief review about these researches. Many image enhancement methods are based on imaging models, and usually described as the problem of inferring intermediate variables with physical sense, e.g., color cast and haze transmission, and then removing them from the input images. Since the problem is ill-posed, it is often solved by enforcing statistical regularities on the intermediate variable. In general, the modeling methods can be divided into two categories: assuming some distributions of the pixels or learning some distributions of pixels from training data. One thing that these assumption or learning based models have in common is that they can all be formulated as the following regression problem: inferring a common variable for a set of candidate pixels (we call it as pixel-ensemble in this paper).

Specifically, for color constancy problem, since the illumination color is usually assumed to be global and consistent, the pixel-ensemble indeed includes all image pixels. Nearly all algorithms for this task work by assuming some distributions of the colors in the pixel-ensemble. For example, the gray world algorithm assumes that the average color of all pixels in intrinsic images is gray. Since then, various methods are proposed to generalize this idea by exploiting gradient information or generalized norms \cite{barnard2002comparison, van2007edge}, modeling the distribution of color histograms \cite{finlayson2001color}, or implicitly reasoning about the moments of colors using PCA \cite{cheng2014illuminant}. Recently, some work further propose to learn the representation of the statistical regularity in the CNN framework \cite{bianco2015color, shi2016deep, hu2017fc}.

For image dehazing problem, the pixel-ensemble can be considered as a local image patch. Based the local constant/smoothness assumption, various methods have been proposed to learn or estimate a transmission value for each local patch \cite{CVPR2009_He, tang2014investigating, cai2016dehazenet}. For example, He et al. propose a powerful dark channel prior to directly estimate haze transmission and then remove the haze from the input image \cite{CVPR2009_He}. Tang et al. investigate four types of haze-relevant features with Random Forests to estimate the transmission \cite{tang2014investigating}. Cai et al. apply a deep CNN framework to regress the transmission from the learning features \cite{cai2016dehazenet}. Recently, Li et al. present an all-in-one CNN to estimate a transformed variable and consequently recover the dehazed image \cite{li2017all}.

In this paper, we propose a pixel shuffling strategy to make the statistical regularity more easily to be detected and more efficiently to be represented by a CNN. Specifically, for a pixel shuffled image, the pixel value tends to be independent identically distributed. It allows us to use an extremely efficient \emph{fully} point-wise CNN to learn the representation of inherent statistical regularity.

\section{Proposed Method}
\subsection{Problem Formulation}
Many inverse problems in image processing can be formulated as follows:
\begin{equation}
y = x + z,
  \label{eq:inverseProblem}
\end{equation}
where $y$ denotes the captured image, $x$ is the underlying ground truth image with which we are concerned, and $z$ is the latent variable which describes noises or other types of influence factors during the imaging process (e.g., color cast or haze transmission). Usually, given the observation $y$, one need to estimate $x$ and $z$, which is an ill-posed inverse problem. Different statistical methods have been proposed by enforcing statistical regularities on the unknown variables \cite{buchsbaum1980spatial,van2007edge,finlayson2004shades,zhu2014single,CVPR2009_He,berman2016non,zhang2017fast} and obtain the estimates by using MAP (Maximum A Posteriori) estimation. Mathematically, it can be formulated as:
\begin{small}
\begin{align}\nonumber
 \left( {{x^*},{z^*}} \right) &= \mathop {\arg \max }\limits_{\left( {x,z} \right)} p\left( {x,z\left| y \right.} \right) \\ \nonumber
  &= \mathop {\arg \max }\limits_{\left( {x,z} \right)} \frac{{p\left( {y\left| {x,z} \right.} \right)p\left( {x,z} \right)}}{{p\left( y \right)}} \\ \nonumber
  &= \mathop {\arg \max }\limits_{\left( {x,z} \right)} \frac{{p\left( {y\left| {x,z} \right.} \right)p\left( x \right)p\left( z \right)}}{{\int\limits_{\rm X} {\int\limits_{\rm Z} {p\left( {y\left| {{x^{'}},{z^{'}}} \right.} \right)p\left( {{x^{'}}} \right)p\left( {{z^{'}}} \right)d{x^{'}}d{z^{'}}} } }} \\
  &= \mathop {\arg \max }\limits_{\left( {x,z} \right)} p\left( {y\left| {x,z} \right.} \right)p\left( x \right)p\left( z \right).
  \label{eq:MAP}
\end{align}
\end{small}
The penultimate equality holds since $x$ and $z$ are usually assumed to be independent. The last equality holds since the denominator is always positive and does not depend on $x$ and $z$. $p\left( {y\left| {x,z} \right.} \right)$ is the data likelihood. $p\left( x \right)$ and $p\left( z \right)$ are prior distributions over $x$ and $z$, respectively.

A common case is to estimate a constant $z$ from an image patch which depends on the statistics of pixels in it: ${y_i} = {x_i} + z,i \in \Omega $. $\Omega$ denotes the index set of pixels in an image patch. If $x_i$ is independently and identically distributed, the data likelihood can be expressed as follows:
\begin{align}\nonumber
 L &= \prod\limits_{i \in \Omega } {p\left( {{y_i}\left| {{x_i},z} \right.} \right)} = \prod\limits_{i \in \Omega } {{p_X}\left( {{y_i} - z} \right)},
  \label{eq:likelihood}
\end{align}
where ${p_X}\left(  \cdot  \right)$ denotes the distribution function of $x$. Therefore, one can estimate $z$ based on ML (Maximum Likelihood) or MAP by incorporating the prior distribution $p\left( z \right)$. Mathematically, the estimate of $z$ can be expressed as a function of the observations:
\begin{equation}
\widehat z = f\left( {{y_1},...,{y_{\left| \Omega  \right|}}} \right),
  \label{eq:mappingFunction}
\end{equation}
where $\left| \Omega  \right|$ denotes the cardinality of $\Omega$.

It can be seen that the explicit form of the mapping $f\left(  \cdot  \right)$ depends on two factors: 1) the prior distribution function ${p_X}\left(  \cdot  \right)$, 2) the estimation method. The goal of this research is to propose a novel method based on deep neural network which can efficiently learn the mapping function $f\left(  \cdot  \right)$ from the observation data by implicitly modelling the statistical regularities in natural images.

\subsection{Motivation}
\label{subsec:motivaiton}
Intuitively, shuffling the pixels in an image does not change their statistics while destroying their spatial structures. It leads to the idea of proposing a novel efficient CNN architecture to learn the statistical regularity from the shuffled image. Since the pixels in the shuffled image tend to be independent identically distributed, we can replace all the large convolution kernels in CNN with point-wise (1*1) convolution kernels while maintaining the representation ability. Moreover, this architecture reduces the risk of network overfitting since it has less parameters and the pixel shuffling strategy eliminates the interference of local structure properties.

\begin{figure}[t]
\begin{center}
\includegraphics[width=1\linewidth]{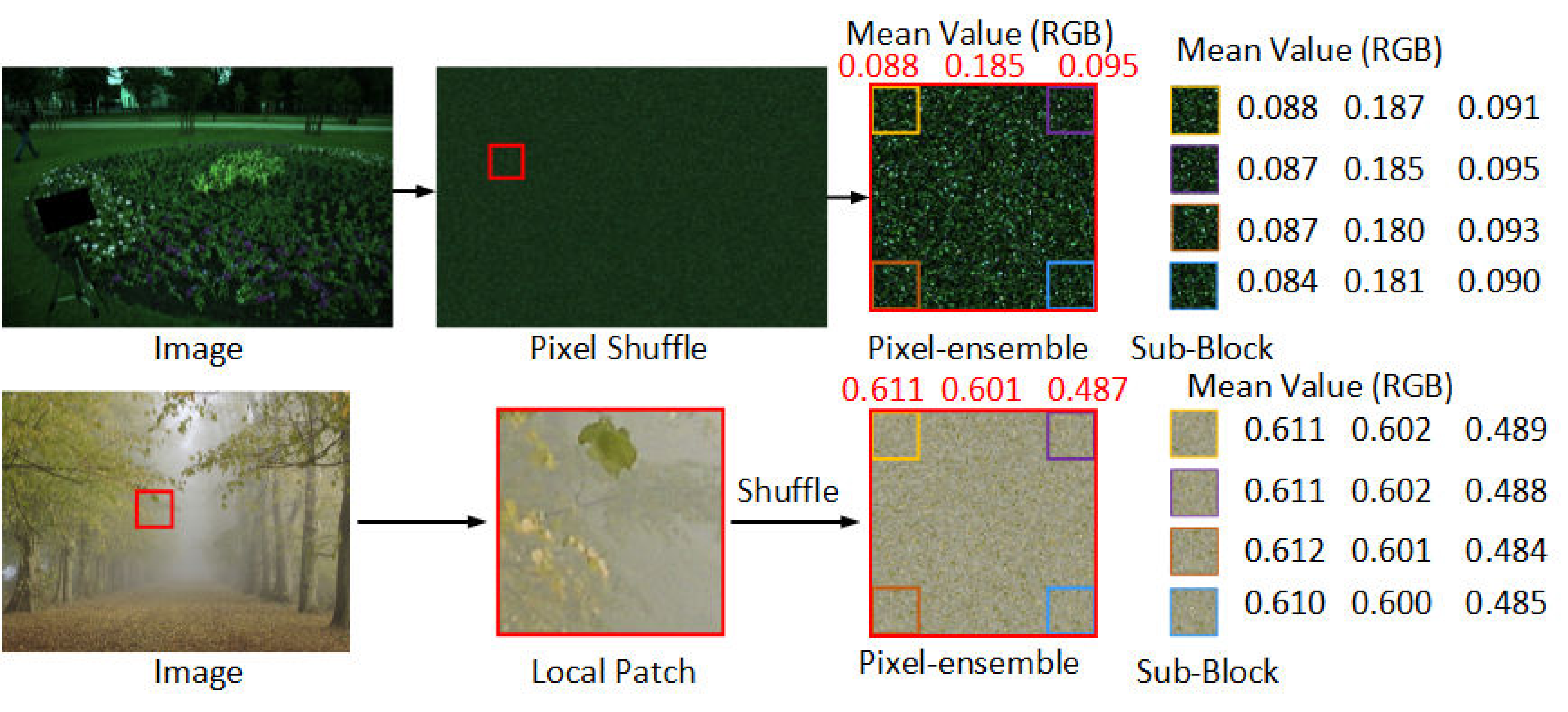}
\end{center}
   \caption{Examples of the pixel-ensemble.}
\label{fig:pixel-ensemble}
\end{figure}

Here we will present a brief proof that we can use point-wise ($1*1$) convolution kernel to replace large ($k*k$) convolution kernel. First, we denote the pixels from one shuffled image as a pixel-ensemble in this paper, i.e.,
\begin{equation}
  X = \left\{ {I(m,n,c),(m,n) \in \Lambda ,c = 0,...C - 1} \right\},
  \label{eq:X}
\end{equation}
where $I(m,n,c)$ is a pixel from the shuffled image, $C$ is the number of image channels and $\Lambda$ is the pixel index set. Pixels in $X$ can be seen as being sampled from the distribution ${p_X}\left(  \cdot  \right)$ independently, i.e., they share the independent and identically distributed property (i.i.d.). Therefore, the mean value of pixels in any subset ${\Lambda _s}$ of $\Lambda$ is approximated to the mean value $\mu_c$ of $X$ on the $c^{th}$ channel, i.e.,
\begin{align}\nonumber
  \mu_c &= \frac{1}{\Lambda}\sum_{(m,n)\in\Lambda}I(m,n,c) \\
  &\approx \frac{1}{{\left| {{\Lambda _s}} \right|}}\sum\limits_{\left( {m,n} \right) \in {\Lambda _s}} {I\left( {m,n,c} \right)},
  \label{eq:muC}
\end{align}
Fig.~\ref{fig:pixel-ensemble} shows two examples for color constancy and image dehazing, respectively. As can be seen, after the pixels of input image/patch are shuffled, the distribution of the generated pixel-ensemble satisfies i.i.d. and the statistics of each sub-block are the same as those of the pixel-ensemble.

Then, let us consider the exemplar network architecture where the input $I$ of shape $(2k-1)*(2k-1)$ is first convolved by a $k \times k \times C$ kernel $K$, and then pooled to be a single value. Without loss of generality, we assume the stride in convolution layer is 1. The output can be calculated as follows:
\begin{small}
\begin{align}
 output &= \frac{1}{{{k^2}}}\sum\limits_{p = 0}^{{k^2} - 1} {\sum\limits_{c = 0}^{C - 1} {\sum\limits_{m,n = 0}^{k - 1} {\left( {{I_p}\left( {m,n,c} \right) \times K\left( {m,n,c} \right)} \right)} } } \nonumber\\
  &= \frac{1}{{{k^2}}}\sum\limits_{c = 0}^{C - 1} {\sum\limits_{m,n = 0}^{k - 1} {\sum\limits_{p = 0}^{{k^2} - 1} {\left( {{I_p}\left( {m,n,c} \right) \times K\left( {m,n,c} \right)} \right)} } }.
\label{eq:output1}
\end{align}
\end{small}Here ${I_p}$ is the $p^{th}$ patch with the size of $k*k$ in the input, and ${I_p}\left( {m,n,c} \right)$ is the pixel located at $(m,n)$ in the $c^{th}$ channel. $\sum\limits_{p = 0}^{p = {k^2} - 1} {{I_p}\left( {m,n,c} \right)} $ is the sum of the $k^2$ pixels in $X$ and approximated as ${k^2}{\mu _c}$ according to Eq.\eqref{eq:muC}. Thus, we have:
\begin{small}
\begin{align}\nonumber
output &= \frac{1}{{{k^2}}}\sum\limits_{c = 0}^{C - 1} {\sum\limits_{m,n = 0}^{k - 1} {\left( {{k^2}{\mu _c} \times K\left( {m,n,c} \right)} \right)} }  \\ \nonumber
  &= \sum\limits_{c = 0}^{C - 1} {{\mu _c}\sum\limits_{m,n = 0}^{k - 1} {K\left( {m,n,c} \right)} }  \\
  &= \sum\limits_{c = 0}^{C - 1} {{\mu _c}\times {K_c}},
\label{eq:output2}
\end{align}
\end{small}
where ${K_c}$ denotes the sum of kernel weights for the $c^{th}$ channel. Substituting Eq.\eqref{eq:muC} into Eq.\eqref{eq:output2}, we have:
\begin{equation}
output = \frac{1}{{{{\left( {2k - 1} \right)}^2}}}\sum\limits_{m,n = 0}^{2k - 2} {\sum\limits_{c = 0}^{C - 1} {I\left( {m,n,c} \right)\times{K_c}} }.
\label{eq:output3}
\end{equation}
As can be seen, given a pixel-ensemble as input, using a large $k*k$ convolution kernel is equivalent to using a $1*1$ convolution kernel, i.e., a network structure with less parameters has the same representation ability with its heavy counterpart. According to the \emph{Occam's razor} principle, a simple model is preferred and resists to overfitting. Therefore, we can design a novel lightweight and efficient network accordingly.

\subsection{Point-wise Convolution Units}
\label{subsec:PointwiseConvUnit}

\begin{figure}[t]
\begin{center}
\includegraphics[width=1\linewidth]{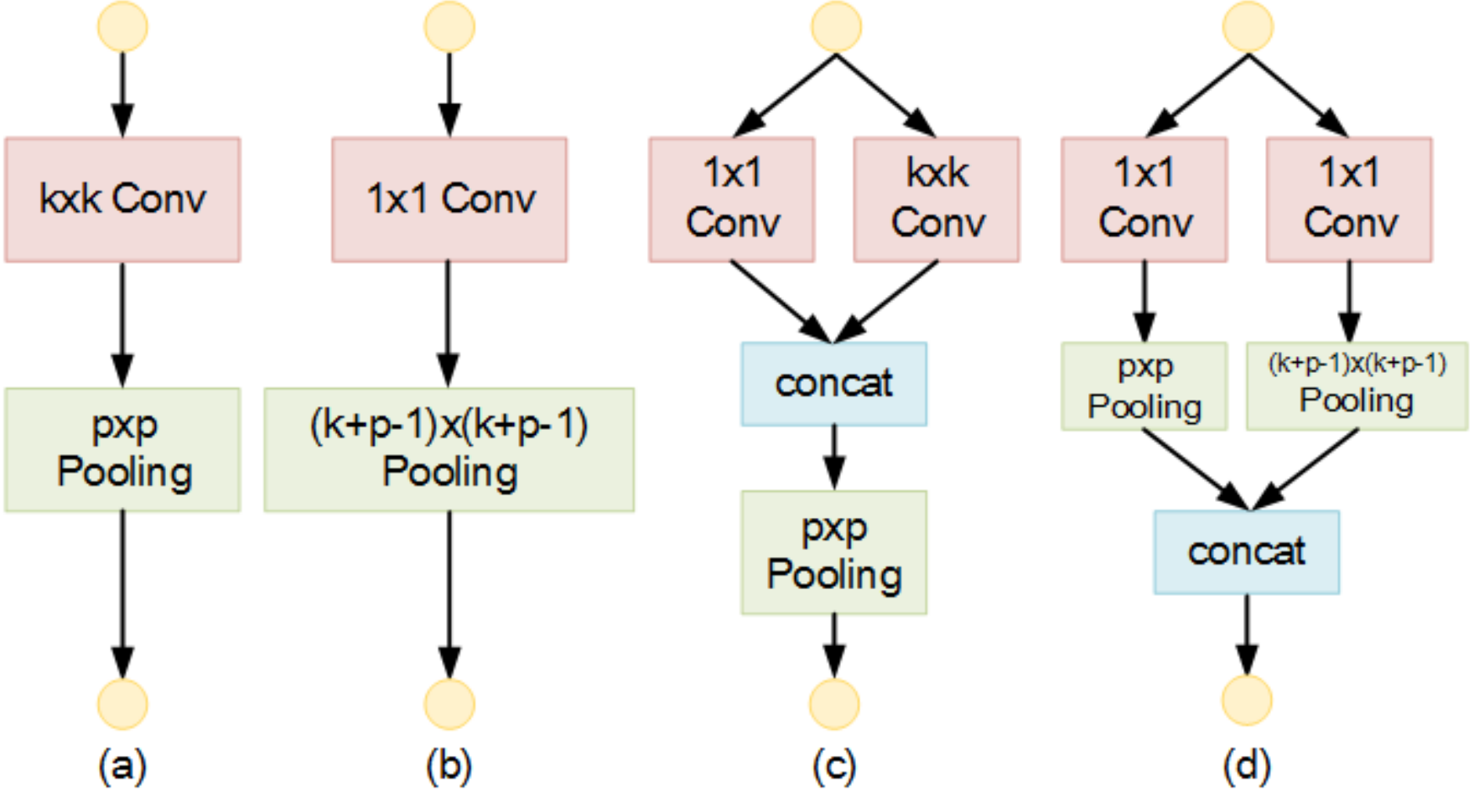}
\end{center}
   \caption{The proposed point-wise convolution units.}
\label{fig:fig4-pointwiseUnit}
\end{figure}

In this part, we propose two novel point-wise convolution units, which can be used to specially design an efficient fully point-wise CNN architecture for modeling statistical regularity. We begin with a typical network structure as shown in Fig.~\ref{fig:fig4-pointwiseUnit}(a), where a convolution layer with $k*k$ convolution kernels is followed by a pooling layer. According to Eq.\eqref{eq:output3}, given the input as a pixel-ensemble, the $k*k$ convolution can be replaced with the point-wise convolution. To retain the size of receptive field, we enlarge the pooling size from $p*p$ to $(k+p-1)*(k+p-1)$, as shown in Fig.~\ref{fig:fig4-pointwiseUnit}(b).

In addition, a parallel structure including a point-wise convolution layer and a $3*3$ convolution layer to extract multi-scale features \cite{szegedy2015going, cai2016dehazenet, ren2016single} is shown in Fig.~\ref{fig:fig4-pointwiseUnit}(c). The extracted features are then concatenated and pooled. Note that the order of the concatenation layer and pooling layer is interchangeable without affecting the result. Similarly, the $k*k$ convolution can be replaced with the point-wise convolution, which results in two parallel point-wise convolution layers, as shown in Fig.~\ref{fig:fig4-pointwiseUnit}(d). And the pooled features from two pooling layers with different pooling sizes are concatenated together.

Taking advantage of the point-wise convolution kernel, our proposed units have less parameters and can be computed efficiently. For example, given the input size $c*h*w$ and the output channels $m$, the unit in Fig.~\ref{fig:fig4-pointwiseUnit}(a) requires $cmk^2$ parameters, while our unit requires only $cm$ parameters. In addition, the point-wise convolution is indeed a scalar-multiplication and an add-operation. Its implementations are more efficiently than the ones with large kernels. By stacking several point-wise convolution units together, we can construct a fully point-wise convolutional neural network (FPCNet). The detail and computational complexity of the explicit network architecture will be presented in Sect.\ref{sec:Applications} since it depends on the specific tasks.

 It is worth noting that though point-wise convolution has been used in several modern deep neural networks such as Inception \cite{szegedy2015going}, ResNet \cite{he2016deep}, MobileNet \cite{howard2017mobilenets} and ShuffleNet \cite{zhang2017shufflenet}, the proposed one is totally different from theirs. Specifically, 1) the proposed architecture is a $FULLY$ point-wise convolutional one without any large convolutional kernels. 2) The proposed architecture utilizing point-wise convolutions in a cascaded manner to aggregate the statistics, while they use it for feature dimension reduction or feature fusion in a single bottleneck layer. 3) The proposed architecture adopts the pixel-ensemble as input which has destroyed the spatial structures of images to model statistical regularities, while they use an intact image as input to learn structural features or other types of high level visual patterns.

\section{Applications}
\label{sec:Applications}
To evaluate the effectiveness of the proposed method, we employ it on two typical image enhancement applications, i.e., color constancy and image dehazing. They resemble some common problems in many image processing tasks, such as HDR compression \cite{fattal2002gradient}, low-light enhancement \cite{guo2017lime}, underwater image enhancement \cite{peng2017underwater} and image defocus \cite{cao2013digital}, which need model the statistical regularities in the whole image or the local patch. For each of the mentioned applications, we perform ablation experiments and contrastive experiments on benchmarks against the state-of-the-art methods. In addition, we also propose a method for visual inspection on what the network has learned about the statistical regularities. All the experiments are conducted on the Nvidia Titan X GPUs, and the proposed networks are implemented in Caffe \cite{jia2014caffe}.

\subsection{Color constancy}
\label{subsec:colorConstancy}

\subsubsection{Problem Formulation and Experiment Settings}

An image captured under color illumination is modeled as follows:
\begin{equation}
{I_c} = {J_c} \times {E_c}, c \in \left\{ {R,G,B} \right\},
\label{eq:CC_model}
\end{equation}
where $J_c$ is the RGB value of reflectance under canonical (often white) illumination and $E_c$ is the color cast. Same to \cite{hu2017fc,barron2017fast}, we treat the color cast $E_c$ to be a global constant and leave the non-uniform cases as the future work. Thus the color constancy problem can be formulated as estimating the color cast $E_c$ given an input image $I_c$ and then using it to recover the reflectance $J_c$. As can be seen that by applying a logarithmic operation on both sides of Eq.\eqref{eq:CC_model}, it shares the same form with Eq.\eqref{eq:inverseProblem}. Hence, we can design a FPCNet to model the statistical regularity on the reflectance (${p_J}\left(  \cdot  \right)$) efficiently and learn an accurate mapping from $I$ to $E$.

The evaluation of our color constancy method is performed on two benchmark datasets, i.e., the reprocessed \cite{shi2000re} Color Checker Dataset \cite{gehler2008bayesian} and the NUS 8-Camera Dataset \cite{cheng2014illuminant}. For Color Checker Dataset, we evaluate the proposed method using a three-fold cross-validation as in \cite{hu2017fc, barron2017fast}. Several standard metrics are reported based on the angular error including mean, median, tri-mean of all the errors, etc. For the NUS 8-Camera Dataset, 8 experiments (three-fold cross-validation for each experiment) on the subset for each camera are conducted, and the geometric mean of each error metric is reported. The angular error between the estimated color cast $({E_{est}})$  and the ground truth $({E_{gt}})$ is calculated as follows:
\begin{equation}
\varepsilon {\rm{ = arccos}}\left( {{{\left( {{E_{est}}\cdot{E_{gt}}} \right)} \mathord{\left/
 {\vphantom {{\left( {{E_{est}}\cdot{E_{gt}}} \right)} {\left( {\left\| {{E_{est}}} \right\|\left\| {{E_{gt}}} \right\|} \right)}}} \right.
 \kern-\nulldelimiterspace} {\left( {\left\| {{E_{est}}} \right\|\left\| {{E_{gt}}} \right\|} \right)}}} \right).
\label{eq:etaError}
\end{equation}
The settings of hyper-parameters during training are the same as \cite{bianco2015color}. If not specified, 128 pixel-ensembles are used for testing and median pooling is applied to the network outputs for the proposed method.

\subsubsection{Ablation Experiments}
Here we present ablation experiments and evaluate the performance of our proposed method. Referring to \cite{bianco2015color}, we present the base network (BaseNet) as shown in Table \ref{tab:FPCnet-CC}. Compared with \cite{bianco2015color}, the BaseNet has a fully CNN structure and three separated prediction sub-nets for RGB channels, respectively. Then, we design a novel fully point-wise CNN according to the proposed units in Sect.\ref{subsec:PointwiseConvUnit} for color constancy (FPCNet-CC). The architectures and their numbers of parameters as well as computational complexity are shown in Table \ref{tab:FPCnet-CC}. We use ReLU and MSE loss in the FPCNet-CC. It is trained in 200,000 iterations with a batch size of 128. The training cycle is about 80 minutes.

\begin{table}[htbp]
\footnotesize
\newcommand{\tabincell}[2]{\begin{tabular}{@{}#1@{}}#2\end{tabular}}
  \centering
  \caption{Network architectures for color constancy.}
    \begin{tabular}{p{1.25cm}p{1.75cm}p{1.25cm}p{0.4cm}p{0.5cm}p{0.25cm}p{0.5cm}}
    \toprule
    Network & Type  & Input Size & Num   & Filter & Pad   & Stride \\
    \midrule
    \multirow{8}[0]{*}{BaseNet} & Conv1-1x1 & 3x32x32 & 240   & 1x1   & 0     & 1 \\
          & Conv1-3x3 & 3x32x32 & 240   & 3x3   & 1     & 1 \\
          & Concat1 & 480x32x32 & -     & -     & -     & - \\
          & Maxpool1 & 480x32x32 & -     & 8x8   & 0     & 8 \\
          & Conv2-(RGB) & 480x4x4 & 40    & 4x4   & 0     & 4 \\
          & Conv3-(RGB) & 40x1x1 & 1     & 1x1   & 0     & 1 \\
          & Params & \multicolumn{5}{c}{9.29x10$^5$}         \\
          & Complexity\footnotemark[1] & \multicolumn{5}{c}{8.29x10$^6$}         \\
    \midrule
    \multirow{10}[0]{*}{FPCNet-CC} & Conv1-1 & 3x32x32 & 240   & 1x1   & 0     & 1 \\
          & Maxpool1-1 & 240x32x32 & -     & 8x8   & 0     & 8 \\
          & Conv1-2 & 3x32x32 & 240   & 1x1   & 0     & 1 \\
          & Maxpool1-2 & 240x32x32 & -     & 10x10 & 1     & 8 \\
          & Concat1 & 480x4x4 & -     & -     & -     & - \\
          & Conv2-(RGB) & 480x4x4 & 80    & 1x1   & 0     & 1 \\
          & Maxpool2 & 480x4x4 & -     & 4x4   & 0     & 4 \\
          & Conv3-(RGB) & 80x1x1 & 1     & 1x1   & 0     & 1 \\
          & Params & \multicolumn{5}{c}{1.17x10$^5$}         \\
          & Complexity & \multicolumn{5}{c}{3.32x10$^6$}         \\
    \bottomrule
    \end{tabular}%
  \label{tab:FPCnet-CC}%
\end{table}%

\footnotetext[1]{Evaluated with FLOPs, i.e. the number of floating-point multiplication-adds.}

\begin{table}[htbp]
\footnotesize
\newcommand{\tabincell}[2]{\begin{tabular}{@{}#1@{}}#2\end{tabular}}
  \centering
  \caption{Results of different networks on Color Checker Dataset \cite{shi2000re}. LP: local patch. PE(A): pixel-ensemble w/o (w/) with augmented data.}
    \begin{tabular}{p{2.5cm}p{0.5cm}p{0.5cm}p{0.5cm}p{0.5cm}p{0.5cm}p{0.75cm}}
    \toprule
    Method & Mean  & Med.  & Tri.  & \tabincell{c}{Best\\ 25\%\\}  & \tabincell{c}{Worst\\ 25\%\\} & \tabincell{c}{95\%\\ Quant.}\\
    \midrule
    BaseNet (LP) & 2.64  & 1.98  & 2.10  & 0.61  & 5.84  & 7.22 \\
    BaseNet (PE) & 2.40  & 1.63  & 1.86  & 0.56  & 5.40  & 6.75 \\
    FPCNet-CC (PE) & 2.22  & 1.51  & 1.69  & 0.45  & 5.12  & 6.85 \\
    FPCNet-CC (PEA) & 2.06  & 1.46  & 1.60  & 0.46  & 4.66  & 5.94 \\
    \bottomrule
    \end{tabular}%
  \label{tab:FPC-CC-Ablation}%
\end{table}%

We perform the ablation experiments on Color Checker Dataset and the results are listed in Table \ref{tab:FPC-CC-Ablation}. As in \cite{bianco2015color} and \cite{shi2016deep}, we first use the local patches as the input of the BaseNet. It achieves a mean angle error of 2.64, which is better than 3.07 in \cite{bianco2015color}. Then we use the pixel-ensemble generated by shuffling the image as the input of the BaseNet, the mean angle error is reduced to 2.40. It verifies that learning the statistics from the shuffled image is more efficient than the original one since it does not need to pay attention to the spatial structure any more. In the next, we use the same pixel-ensemble as the input for the proposed FPCNet-CC, the mean angle error is reduced to 2.22, with only 12.5\% network parameters and 40\% computational cost of BaseNet. It demonstrates that our proposed structure models the global statistics more efficiently and learns the mapping function more accurately than its counterpart, i.e., BaseNet. After using the edge pixels as the augmented data (Please refer the fourth equation of Eq.15 in \cite{barron2015convolutional}), the mean angle error is further reduced to 2.06, which is close to the best result achieved by single network methods. Moreover, the results in terms of Worst-25\% metric are significantly improved, which implies that the proposed method achieves a better performance on the hard samples.

\begin{figure}[t]
\begin{center}
\includegraphics[width=1\linewidth]{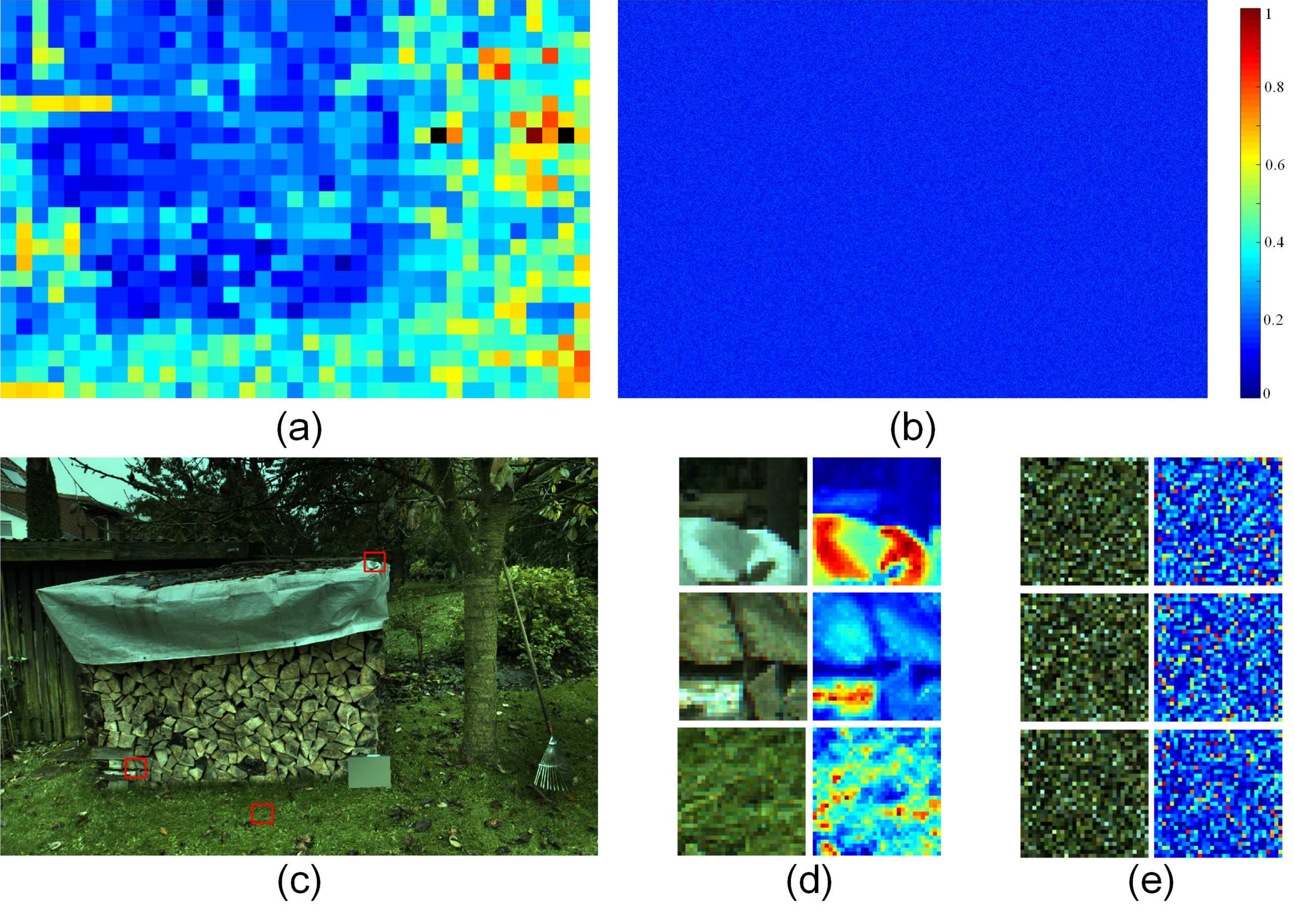}
\end{center}
   \caption{(a)-(b) Error maps of BaseNet and FPCNet-CC. (c) The test image. (d) Examplar local patches and their response maps ($Concat1$) of BaseNet. (e) Examplar pixel-ensembles and their response maps ($Conv1$) of FPCNet-CC .}
\label{fig:fig5-pixelErrorStats}
\end{figure}

To further demonstrate the modeling capacity of each structure, we present the corresponding error maps of BaseNet and FPCNet-CC for a given test image in Fig.~\ref{fig:fig5-pixelErrorStats}(a)-(c). The angle errors of BaseNet are diverse while the angle errors of FPCNet-CC are uniform and small. To shed light on the difference, the response maps after the first level of convolutions in BaseNet and FPCNet-CC are extracted and shown in Fig.~\ref{fig:fig5-pixelErrorStats}(d) and (e), respectively. As can be seen, pixels which have strong activations in BaseNet differ from patch to patch, and are affected by the non-uniform local statistics. As a contrast, FPCNet-CC can handle it well by using pixel-ensembles which show similar global statistics.

\begin{table}[htbp]
\tiny
\newcommand{\tabincell}[2]{\begin{tabular}{@{}#1@{}}#2\end{tabular}}
  \scriptsize
  \caption{Results on NUS 8-Camera Dataset \cite{cheng2014illuminant}.}
    \begin{tabular}{p{2.75cm}p{0.55cm}p{0.55cm}p{0.3cm}p{0.3cm}p{0.3cm}p{0.3cm}p{0.3cm}p{0.5cm}}
    \toprule
    Method & Params & Time(s) & Mean  & Med.  & Tri.  & \tabincell{c}{Best\\ 25\%\\}  & \tabincell{c}{Worst\\ 25\%\\} & \tabincell{c}{95\%\\ Quant.}\\
    \midrule
    White-Patch \cite{brainard1986analysis}& - & 0.16 & 10.62 & 10.58 & 10.49 & 1.86  & 19.45 & 8.43 \\
    Edge-based Gamut \cite{barnard2000improvements}& - & 3.6 & 4.40  & 3.30  & 3.45  & 0.99  & 9.83 & - \\
    Gray-World \cite{buchsbaum1980spatial}& - & 0.15 & 4.14  & 3.20  & 3.39  & 0.90  & 9.00  & 3.25 \\
    Bayesian \cite{gehler2008bayesian}& - & 97 & 3.67  & 2.73  & 2.91  & 0.82  & 8.21  & 2.88 \\
    Natural Image Stat. \cite{gijsenij2011color}& - & 1.5 & 3.71  & 2.60  & 2.84  & 0.79  & 8.47  & 2.83 \\
    Shades-of-Gray \cite{finlayson2004shades}& - & 0.47 & 3.40  & 2.57  & 2.73  & 0.77  & 7.41  & 2.67 \\
    General Gray-World \cite{barnard2002comparison}& - & 0.91 & 3.21  & 2.38  & 2.53  & 0.71  & 7.10  & 2.49 \\
    1st-order Gray-Edge \cite{van2007edge}& - & 1.1 & 3.20  & 2.22  & 2.43  & 0.72  & 7.36  & 2.46 \\
    Bright Pixels \cite{joze2012role}& - & - & 3.17  & 2.41  & 2.55  & 0.69  & 7.02  & 2.48 \\
    Cheng et al. \cite{cheng2014illuminant}& - & 0.24 & 2.92  & 2.04  & 2.24  & 0.62  & 6.61  & 2.23 \\
    CCC(dist+ext) \cite{barron2015convolutional}& - & 0.52 & 2.38  & 1.48  & 1.69  & 0.45  & 5.85  & 1.74 \\
    Regression Tree \cite{cheng2015effective}& - & 0.25 & 2.36  & 1.59  & 1.74  & 0.49  & 5.54  & 1.78 \\
    FFCC-full \cite{barron2017fast}& - & 0.07 & 1.99  & 1.31  & 1.43  & 0.35  & 4.75  & 1.44 \\
    FFCC-thumb \cite{barron2017fast}& - & 0.0011 & 2.06  & 1.39  & 1.43  & 0.35  & 4.75  & 1.44 \\
    \midrule
    CNN \cite{bianco2017single}& 0.154M & 0.208 & - & 1.73 & - & - & - & -\\
    DS-Net \cite{shi2016deep}&4.23M & 3.0 & 2.24  & 1.46  & 1.68  & 0.48  & 5.28  & 1.69 \\
    AlexNet-FC4 \cite{hu2017fc}& 2.48M & 0.025 & 2.12  & 1.53  & 1.67  & 0.48  & 4.78  & 1.66 \\
    SqueezeNet-FC4 \cite{hu2017fc}& 2.12M & 0.025 & 2.23  & 1.57  & 1.72  & 0.47  & 5.15  & 1.71 \\
    \midrule
    FPCNet-CC & 0.117M & 0.0027 & 2.17  & 1.57  & 1.66  & 0.51  & 4.88  & 1.70 \\
    \bottomrule
    \end{tabular}%
  \label{tab:NUS}%
\end{table}%

\subsubsection{Comparisons with State-of-the-art Methods}
Here we compare the proposed FPCNet-CC with previous methods on the NUS 8-Camera Dataset. We report each error metric for each method across all cameras as done in previous work. Results are summarized in Table \ref{tab:NUS}. For most metrics, FPCNet-CC achieves comparable results with the state-of-the-art methods, e.g. FC4 \cite{hu2017fc} and FFCC \cite{barron2017fast}. Compared with the classical statistical prior based method, e.g., Gray-World \cite{buchsbaum1980spatial} and 1st-order Gray-Edge \cite{van2007edge}, FPCNet-CC outperforms them with a significant margin. It verifies that our proposed network which directly learns the statistics from data can model the statistical regularity more effectively than different kinds of ad-hoc statistical priors. Compared with learning based methods, especially CNN based methods, FPCNet-CC achieves comparable or better results with less parameters. Therefore, the proposed FPCNet-CC is promising to serve as an alternative light-weight solution for color constancy.

Moreover, we also compare the computational efficiency with the previous methods. The results in terms of running time are listed in Table \ref{tab:NUS}. Our proposed FPCNet-CC is found to be $10$x $\sim100$x faster than previous CNN based methods \cite{bianco2015color, shi2016deep, hu2017fc}. It processes a single image in 2.7ms, compared to 208ms for \cite{bianco2015color, bianco2017single}, 3s for \cite{shi2016deep}, and 25ms for \cite{hu2017fc}. This advantage is due to the fully point-wise convolutional structure, which can be implemented efficiently. Besides, as shown in Fig.~\ref{fig:fig5-pixelErrorStats}(d), the angle error map is uniform. It implies that we can use less pixel-ensembles instead of 128 to boost the computational efficiency by sacrificing a little accuracy.

\subsubsection{Visual Inspection on the Learned Statistical Regularities}
\label{subsubsec:visualInspection}

\begin{figure}[t]
\begin{center}
\includegraphics[width=0.9\linewidth]{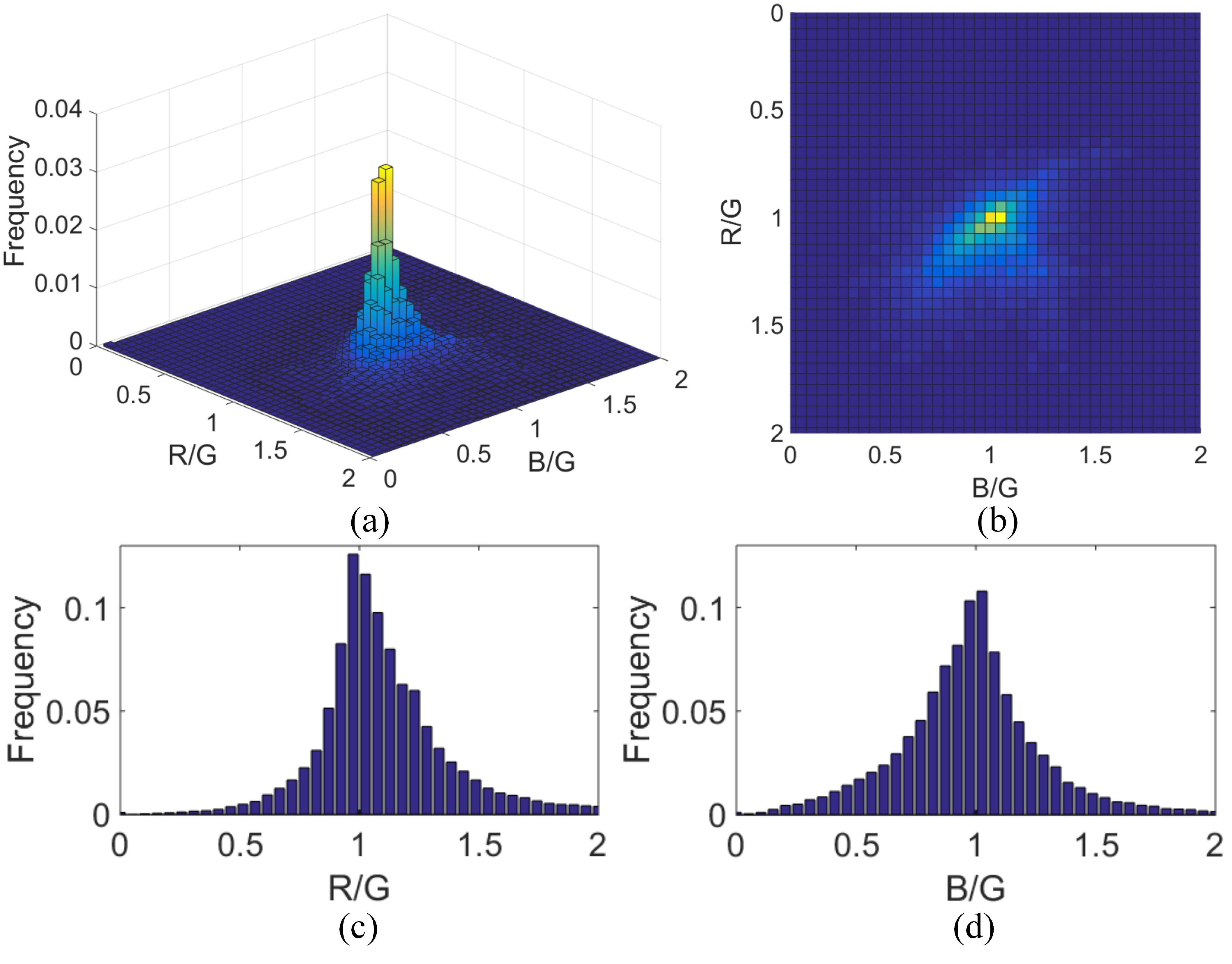}
\end{center}
   \caption{(a)-(b) Weighted histogram of FPCNet-CC and its bird view. (c)-(d) Marginal histograms of (a).}
\label{fig:fig-histogramCC}
\end{figure}

To visually inspect what the proposed FPCNet-CC has learned about the statistical regularities, we propose a weighted histogram method based on neuron activations. First, we calculate the neuron activations of the first pooling layer by averaging the response maps across the feature channels. The neuron activation reflects the importance of each pixel contributing to the final prediction. Then, we re-project them onto the the original image according to the shuffle indexes of pixel-ensemble (The FPCNet has a good backtrace ability due to the point-wise convolution). One example is shown in Fig.~\ref{fig:fig1-illustration}(c), where the pixel values are multiplied by the neuron activations. As can be seen, the most activated pixels include the bright pixels, white pixels, etc, which contain cues of the color cast. By re-projecting them onto the intrinsic image, we hope that those strongly activated pixels have good statistical property. Therefore, for all the re-projected intrinsic images in Color Checker Dataset, we calculate the weighted 2-dimension histogram by counting the accumulated weights(i.e., neuron activations) in each R/G and B/G grid. Ideally, the histogram should concentrate at the center (1,1) with small variance, such that one can estimate the color cast from the ratios between R (and B) channel to G channel according to Eq.\eqref{eq:CC_model}. The histograms of the proposed FPCNet-CC and its bird view are shown in Fig.\ref{fig:fig-histogramCC}(a)-(b). As can be seen, it concentrates on the center (1,1). Moreover, from the marginal histogram in Fig.\ref{fig:fig-histogramCC}(c)-(d), it is clear that the proposed FPCNet-CC is immune to the inherent distribution bias by sampling important pixels to form a more effective statistical regularity, which enables to estimate the color cast easily.

\subsection{Image dehazing}
\label{subsec:imageDehazing}

\subsubsection{Problem Formulation and Experiment Settings}
The formation of a hazy image can be described as follows:
\begin{equation}
{I_c} = {J_c}t + {A_c}\left( {1 - t} \right), c \in \left\{ {R,G,B} \right\},
\label{eq:hazyModel}
\end{equation}
where $J_c$ is the target clear image, $t$ is the haze transmission, $A_c$ is the atmosphere light. Under the local constant/smoothness assumption about $t$, the image dehazing problem can be formulated as estimating a haze transmission $t$ at each local patch given an input hazy image $I$ and using it to recover the clear image $J$. Usually, the atmosphere light $A_c$ is a global constant, thus we can rewrite Eq.\eqref{eq:hazyModel} as:
\begin{equation}
\left( {{I_c} - {A_c}} \right) = \left( {{J_c} - {A_c}} \right) \times t.
\label{eq:hazyModel2}
\end{equation}
It resembles the color constancy model (Eq.\eqref{eq:CC_model}). Hence, we can design a FPCNet to model the statistical regularity on the the underlying clear image (${p_J}\left(  \cdot  \right)$) efficiently and learn an accurate mapping function from $I$ to $t$.

Since hazy images with ground truth transmissions are hard to collect, we build the synthesized hazy image dataset as in \cite{cai2016dehazenet}. First, we collect 250 clear
images and split them into non-overlapped train/test sets (200/50). Then, 30,000 haze-free patches are randomly sampled from them. A total of 300, 000 synthetic hazy image patches are generated according to Eq.\eqref{eq:hazyModel}. Samples in the corresponding splits are used for training and testing, respectively.

\begin{table}[htbp]
\footnotesize
  \newcommand{\tabincell}[2]{\begin{tabular}{@{}#1@{}}#2\end{tabular}}
  \centering
  \caption{Network architectures for dehazing.}
    \begin{tabular}{p{1.5cm}p{1.5cm}p{1.25cm}p{0.4cm}p{0.5cm}p{0.25cm}p{0.5cm}}
    \toprule
    Network & Type  & Input Size & Num   & Filter & Pad   & Stride \\
    \midrule
    \multirow{8}[0]{*}{FPCNet-DH} & Conv1 & 3x16x16 & 16    & 1x1   & 0     & 1 \\
          & Maxout & 16x16x16 & -     & 4x1   & -     & - \\
          & Maxpool1 & 4x16x16 & -     & 2x2   & 0     & 2 \\
          & Conv2 & 4x8x8 & 48    & 1x1   & 0     & 1 \\
          & Maxpool2 & 48x8x8 & -     & 8x8   & 0     & 8 \\
          & Conv3 & 48x1x1 & 1     & 1x1   & 0     & 1 \\
          & Params & \multicolumn{5}{c}{288}               \\
          & Complexity & \multicolumn{5}{c}{2.46x10\tiny{$^4$}}        \\
    \midrule
    \multirow{2}[0]{*}{DehazeNet} & Params & \multicolumn{5}{c}{8240}              \\
          & Complexity & \multicolumn{5}{c}{9.39x10$^5$}        \\
    \bottomrule
    \end{tabular}%
  \label{tab:FPC-Dehaze}%
\end{table}%

\subsubsection{Ablation Experiments}
\label{subsubsec:dehazingAblation}
Here we present ablation experiments to design the network for image dehazing. We refer dehazenet \cite{cai2016dehazenet} as our baseline and present a novel fully point-wise CNN according to the proposed units in Sect.~\ref{subsec:PointwiseConvUnit} for dehazing (FPCNet-DH). We insert a pooling layer before the multi-scale feature layers with stride 2 to reduce the computational complexity. Since max pooling is more efficient in preserving useful features than average pooling in practice, we use max pooling as our default setting. The architectures and their numbers of parameters as well as computational complexity are shown in Table \ref{tab:FPC-Dehaze}. We use BReLU after the last convolutional layer and MSE loss in the FPCNet-DH. The settings of hyper-parameters are the same as \cite{cai2016dehazenet}.

To verify the modeling capacity of the proposed FPCNet-DH, we compare it with state-of-the-art methods according to the MSE of predicted transmission. As shown in Table \ref{tab:tMSE}, FPCNet-DH achieves much better results than DCP\cite{CVPR2009_He}. Compared with the state-of-the-art (DehazeNet), the proposed FPCNet-DH achieves comparable results by using 3.5\% network parameters and 2.62\% computational cost. Then, we test the effectiveness of the proposed method on complete synthesized images. We synthesize hazy images based on the stereo images from Middlebury Stereo Datasets \cite{hirschmuller2007evaluation} by referring \cite{cai2016dehazenet}. The PSNR and SSIM results of different methods are summarized in Table \ref{tab:middleburyPsnrSsim}. The proposed FPCNet-DH achieves better results than DCP and Dehazenet.

\begin{table}[htbp]
  \centering
  \caption{MSE of the predicted transmission on the test set of different methods.}
    \begin{tabular}{lccc}
    \toprule
    Methods & DCP\cite{CVPR2009_He}   & DehazeNet\cite{cai2016dehazenet} & FPCNet-DH\\
    \midrule
    MSE(x10$^{-2}$) & 2.41  & 1.20  & 1.17 \\
    \bottomrule
    \end{tabular}%
  \label{tab:tMSE}%
\end{table}%

\begin{table}[htbp]
  \centering
  \caption{PSNR and SSIM of dehazing results on complete synthetic hazy images of different methods.}
    \begin{tabular}{lccc}
     \toprule
    Methods & DCP\cite{CVPR2009_He}   & DehazeNet\cite{cai2016dehazenet} & FPCNet-DH \\
     \midrule
    PSNR  & 20.16 & 20.29 & 21.17 \\
     \hline
    SSIM  & 0.8611 & 0.8680 & 0.8733 \\
     \bottomrule
    \end{tabular}%
  \label{tab:middleburyPsnrSsim}%
\end{table}%

\subsubsection{Comparisons with State-of-the-art Methods}
\label{subsubsec:dehazingComp}
We compare the proposed method with state-of-the-art methods including DCP \cite{CVPR2009_He}, dehazeNet \cite{cai2016dehazenet} and AODNet \cite{li2017all} on outdoor real hazy images. Codes and models of these methods are provided by the authors. Figure~\ref{fig:fig6-dehazeResults} shows the visual inspection results on some challenging natural hazy images. As can be seen, our proposed FPCNet-DH achieves the most competitive visual results among all. It demonstrates the proposed FPCNet-DH has superior capacity of modeling the statistics in natural images, even for the challenging cases, such as the illumination variant regions and textureless regions.

To compare the running time of different methods, we test them on the hazy images with a size of $640*480$. All the networks are tested on Matlab platform with GPU acceleration. The proposed FPCNet-DH processes a single image in 3ms, while DehazeNet takes 466ms and AODNet takes 4.3ms. The proposed FPCNet-DH is the fastest due to its light-weight and fully point-wise convolutional structure.

\begin{figure}[t]
\begin{center}
\includegraphics[width=1\linewidth]{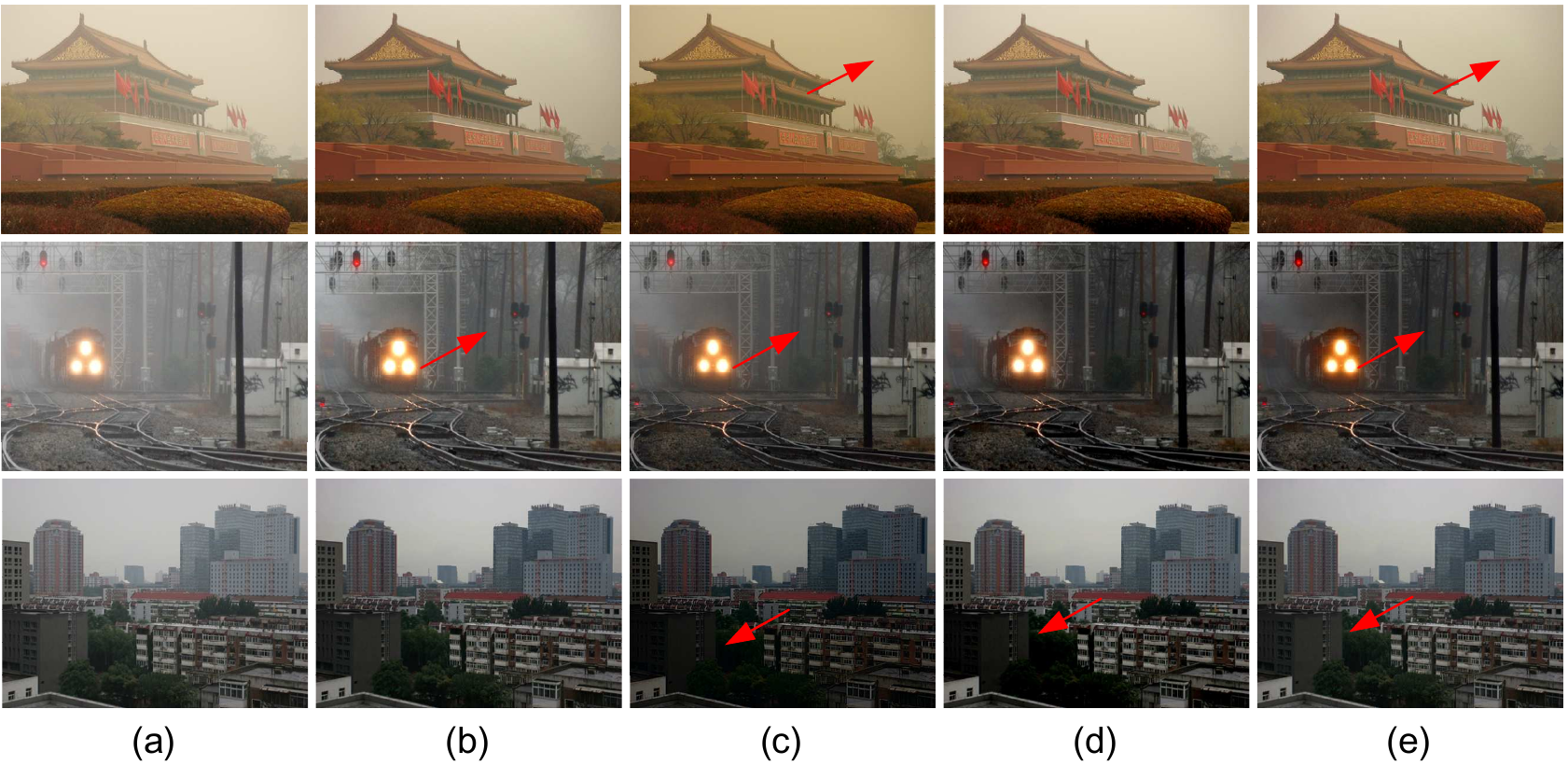}
\end{center}
   \caption{(a)Hazy images. (b)Results of DCP \cite{CVPR2009_He}. (c)Results of AODNet \cite{li2017all}. (d)Results of DehazeNet \cite{cai2016dehazenet}. (e)Results of the proposed FPCNet-DH.}
\label{fig:fig6-dehazeResults}
\end{figure}

\subsubsection{Visual Inspection on the Learned Statistical Regularities}
We use the method in Sect.~\ref{subsubsec:visualInspection} to calculate the weighted histogram according to the neuron activations of the second max-pooling layer. Specifically, here we concern with the histogram of the minimal values of the $R$, $G$ and $B$ channels, since it has a close relation to the transmission (By applying a minimum operation across channels on both sides of Eq.\eqref{eq:hazyModel2}, one can easily obtain a mapping from $I$ to $t$ \cite{CVPR2009_He}). We calculate the histograms on 100 clear images both for FPCNet-DH and DCP \cite{CVPR2009_He}. Results are shown in Fig.~\ref{fig:fig6-dehazeResults}. As can be seen, histogram of FPCNet-DH leans towards to small values more heavily than the one of DCP, and the corresponding cumulative distribution goes up much faster. In other word, the proposed FPCNet-DH forms a more effective statistical regularity by sampling important pixels. Consequently, it enables to learn a more accurate mapping from the hazy image to the haze transmission.

\begin{figure}[t]
\begin{center}
\includegraphics[width=1\linewidth]{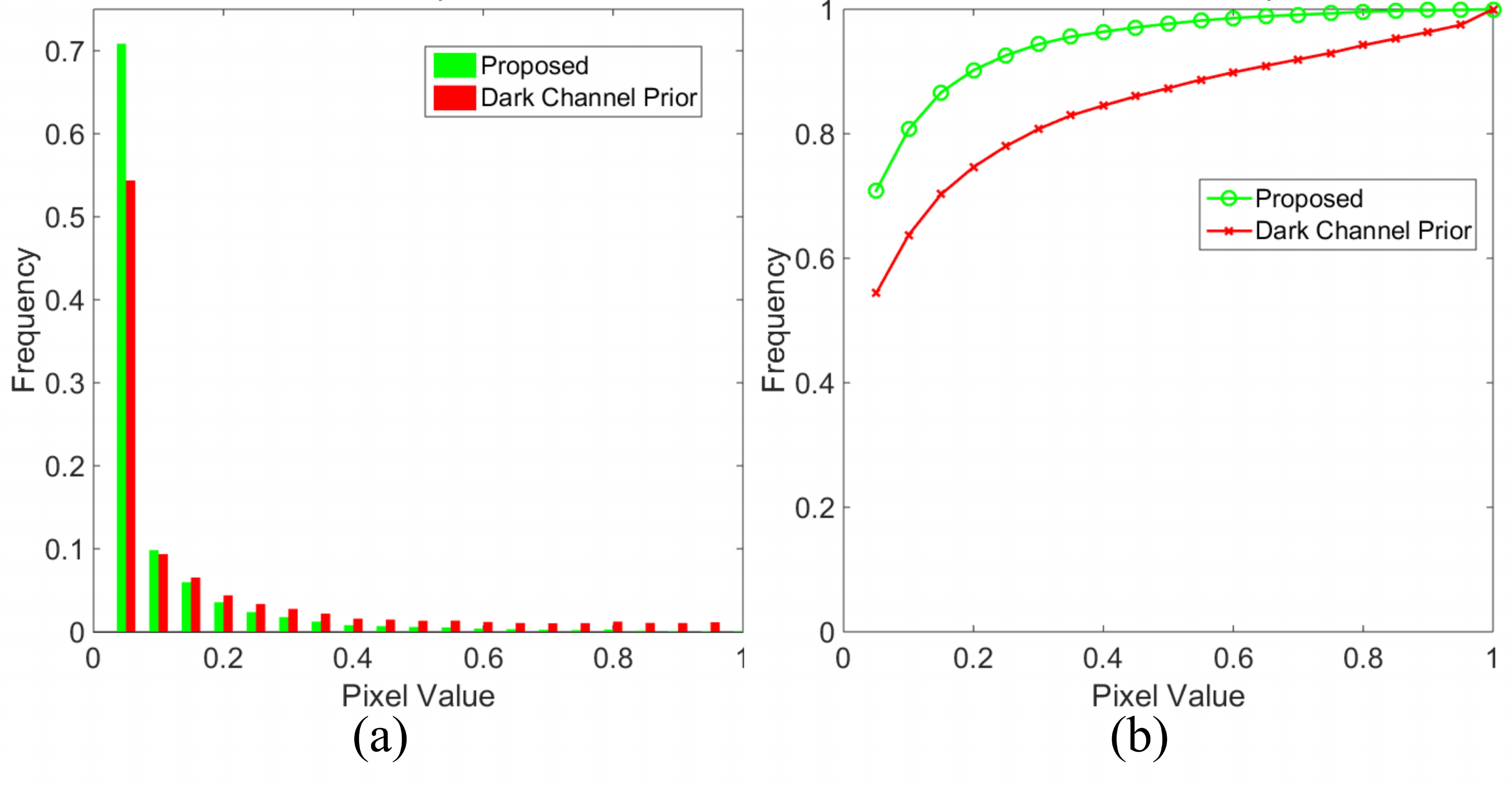}
\end{center}
   \caption{(a)-(b)Histograms and cumulative distributions of DCP \cite{CVPR2009_He} and the proposed FPCNet-DH.}
\label{fig:fig6-dehazeResults}
\end{figure}

\section{Conclusion}
In this paper, we introduce a fully point-wise CNN (FPCNet) method which uses point-wise convolutions in all convolutional layers instead of any large kernels. By using a pixel-ensemble as input which is generated by shuffling the original image, the proposed FPCNet can model the statistical regularities effectively. The comprehensive evaluations on color constancy and image dehazing demonstrate that our proposed FPCNet achieves the superior efficiency over the existing architectures while maintaining comparable accuracy. It is promising to be an alternative and complementary statistical method for solving various ill-posed problems.

The limitation of our proposed method is that it only captures the statistical distribution while misses the spatial structures in the pixel space. One feasible solution for this is to design a multi-branch network and take our proposed structures as one of them. Another solution is to connect our proposed structures to the network that extracts structural features, which directly models the statistical properties in the feature space. Due to the lightweight and compact properties, our proposed architecture is promising to work well in the two scenarios. We will it as the future work.

\section*{Acknowledgment}
This work was supported by the National Natural Science Foundation of China (NSFC) under Grants 61806062, 61472380, 61751304 and 61873077.

\bibliographystyle{ACM-Reference-Format}

\begin{thebibliography}{43}


\ifx \showCODEN    \undefined \def \showCODEN     #1{\unskip}     \fi
\ifx \showDOI      \undefined \def \showDOI       #1{#1}\fi
\ifx \showISBNx    \undefined \def \showISBNx     #1{\unskip}     \fi
\ifx \showISBNxiii \undefined \def \showISBNxiii  #1{\unskip}     \fi
\ifx \showISSN     \undefined \def \showISSN      #1{\unskip}     \fi
\ifx \showLCCN     \undefined \def \showLCCN      #1{\unskip}     \fi
\ifx \shownote     \undefined \def \shownote      #1{#1}          \fi
\ifx \showarticletitle \undefined \def \showarticletitle #1{#1}   \fi
\ifx \showURL      \undefined \def \showURL       {\relax}        \fi
\providecommand\bibfield[2]{#2}
\providecommand\bibinfo[2]{#2}
\providecommand\natexlab[1]{#1}
\providecommand\showeprint[2][]{arXiv:#2}

\bibitem[\protect\citeauthoryear{Barnard}{Barnard}{2000}]%
        {barnard2000improvements}
\bibfield{author}{\bibinfo{person}{Kobus Barnard}.}
  \bibinfo{year}{2000}\natexlab{}.
\newblock \showarticletitle{Improvements to gamut mapping colour constancy
  algorithms}.
\newblock \bibinfo{journal}{\emph{Computer Vision-ECCV 2000}}
  (\bibinfo{year}{2000}), \bibinfo{pages}{390--403}.
\newblock


\bibitem[\protect\citeauthoryear{Barnard, Cardei, and Funt}{Barnard
  et~al\mbox{.}}{2002}]%
        {barnard2002comparison}
\bibfield{author}{\bibinfo{person}{Kobus Barnard}, \bibinfo{person}{Vlad
  Cardei}, {and} \bibinfo{person}{Brian Funt}.}
  \bibinfo{year}{2002}\natexlab{}.
\newblock \showarticletitle{A comparison of computational color constancy
  algorithms. I: Methodology and experiments with synthesized data}.
\newblock \bibinfo{journal}{\emph{IEEE transactions on Image Processing}}
  \bibinfo{volume}{11}, \bibinfo{number}{9} (\bibinfo{year}{2002}),
  \bibinfo{pages}{972--984}.
\newblock


\bibitem[\protect\citeauthoryear{Barron}{Barron}{2015}]%
        {barron2015convolutional}
\bibfield{author}{\bibinfo{person}{Jonathan~T Barron}.}
  \bibinfo{year}{2015}\natexlab{}.
\newblock \showarticletitle{Convolutional color constancy}. In
  \bibinfo{booktitle}{\emph{Proceedings of the IEEE International Conference on
  Computer Vision}}. \bibinfo{pages}{379--387}.
\newblock


\bibitem[\protect\citeauthoryear{Barron and Tsai}{Barron and Tsai}{2017}]%
        {barron2017fast}
\bibfield{author}{\bibinfo{person}{Jonathan~T Barron} {and}
  \bibinfo{person}{Yun-Ta Tsai}.} \bibinfo{year}{2017}\natexlab{}.
\newblock \showarticletitle{Fast Fourier Color Constancy}.
\newblock \bibinfo{journal}{\emph{Proceedings of the IEEE Conference on
  Computer Vision and Pattern Recognition}} (\bibinfo{year}{2017}).
\newblock


\bibitem[\protect\citeauthoryear{Berman, Avidan, et~al\mbox{.}}{Berman
  et~al\mbox{.}}{2016}]%
        {berman2016non}
\bibfield{author}{\bibinfo{person}{Dana Berman}, \bibinfo{person}{Shai Avidan},
  {et~al\mbox{.}}} \bibinfo{year}{2016}\natexlab{}.
\newblock \showarticletitle{Non-local image dehazing}. In
  \bibinfo{booktitle}{\emph{Proceedings of the IEEE conference on computer
  vision and pattern recognition}}. \bibinfo{pages}{1674--1682}.
\newblock


\bibitem[\protect\citeauthoryear{Bianco, Cusano, and Schettini}{Bianco
  et~al\mbox{.}}{2015}]%
        {bianco2015color}
\bibfield{author}{\bibinfo{person}{Simone Bianco}, \bibinfo{person}{Claudio
  Cusano}, {and} \bibinfo{person}{Raimondo Schettini}.}
  \bibinfo{year}{2015}\natexlab{}.
\newblock \showarticletitle{Color constancy using CNNs}. In
  \bibinfo{booktitle}{\emph{Proceedings of the IEEE Conference on Computer
  Vision and Pattern Recognition Workshops}}. \bibinfo{pages}{81--89}.
\newblock


\bibitem[\protect\citeauthoryear{Bianco, Cusano, and Schettini}{Bianco
  et~al\mbox{.}}{2017}]%
        {bianco2017single}
\bibfield{author}{\bibinfo{person}{Simone Bianco}, \bibinfo{person}{Claudio
  Cusano}, {and} \bibinfo{person}{Raimondo Schettini}.}
  \bibinfo{year}{2017}\natexlab{}.
\newblock \showarticletitle{Single and Multiple Illuminant Estimation Using
  Convolutional Neural Networks}.
\newblock \bibinfo{journal}{\emph{IEEE Transactions on Image Processing}}
  (\bibinfo{year}{2017}).
\newblock


\bibitem[\protect\citeauthoryear{Brainard and Wandell}{Brainard and
  Wandell}{1986}]%
        {brainard1986analysis}
\bibfield{author}{\bibinfo{person}{David~H Brainard} {and}
  \bibinfo{person}{Brian~A Wandell}.} \bibinfo{year}{1986}\natexlab{}.
\newblock \showarticletitle{Analysis of the retinex theory of color vision}.
\newblock \bibinfo{journal}{\emph{JOSA A}} \bibinfo{volume}{3},
  \bibinfo{number}{10} (\bibinfo{year}{1986}), \bibinfo{pages}{1651--1661}.
\newblock


\bibitem[\protect\citeauthoryear{Buchsbaum}{Buchsbaum}{1980}]%
        {buchsbaum1980spatial}
\bibfield{author}{\bibinfo{person}{Gershon Buchsbaum}.}
  \bibinfo{year}{1980}\natexlab{}.
\newblock \showarticletitle{A spatial processor model for object colour
  perception}.
\newblock \bibinfo{journal}{\emph{Journal of the Franklin institute}}
  \bibinfo{volume}{310}, \bibinfo{number}{1} (\bibinfo{year}{1980}),
  \bibinfo{pages}{1--26}.
\newblock


\bibitem[\protect\citeauthoryear{Burger, Schuler, and Harmeling}{Burger
  et~al\mbox{.}}{2012}]%
        {burger2012image}
\bibfield{author}{\bibinfo{person}{Harold~C Burger},
  \bibinfo{person}{Christian~J Schuler}, {and} \bibinfo{person}{Stefan
  Harmeling}.} \bibinfo{year}{2012}\natexlab{}.
\newblock \showarticletitle{Image denoising: Can plain neural networks compete
  with BM3D?}. In \bibinfo{booktitle}{\emph{Computer Vision and Pattern
  Recognition (CVPR), 2012 IEEE Conference on}}. IEEE,
  \bibinfo{pages}{2392--2399}.
\newblock


\bibitem[\protect\citeauthoryear{Cai, Xu, Jia, Qing, and Tao}{Cai
  et~al\mbox{.}}{2016}]%
        {cai2016dehazenet}
\bibfield{author}{\bibinfo{person}{Bolun Cai}, \bibinfo{person}{Xiangmin Xu},
  \bibinfo{person}{Kui Jia}, \bibinfo{person}{Chunmei Qing}, {and}
  \bibinfo{person}{Dacheng Tao}.} \bibinfo{year}{2016}\natexlab{}.
\newblock \showarticletitle{Dehazenet: An end-to-end system for single image
  haze removal}.
\newblock \bibinfo{journal}{\emph{IEEE Transactions on Image Processing}}
  \bibinfo{volume}{25}, \bibinfo{number}{11} (\bibinfo{year}{2016}),
  \bibinfo{pages}{5187--5198}.
\newblock


\bibitem[\protect\citeauthoryear{Cao, Fang, and Wang}{Cao
  et~al\mbox{.}}{2013}]%
        {cao2013digital}
\bibfield{author}{\bibinfo{person}{Yang Cao}, \bibinfo{person}{Shuai Fang},
  {and} \bibinfo{person}{Zengfu Wang}.} \bibinfo{year}{2013}\natexlab{}.
\newblock \showarticletitle{Digital multi-focusing from a single photograph
  taken with an uncalibrated conventional camera}.
\newblock \bibinfo{journal}{\emph{IEEE Transactions on image processing}}
  \bibinfo{volume}{22}, \bibinfo{number}{9} (\bibinfo{year}{2013}),
  \bibinfo{pages}{3703--3714}.
\newblock


\bibitem[\protect\citeauthoryear{Cheng, Prasad, and Brown}{Cheng
  et~al\mbox{.}}{2014}]%
        {cheng2014illuminant}
\bibfield{author}{\bibinfo{person}{Dongliang Cheng}, \bibinfo{person}{Dilip~K
  Prasad}, {and} \bibinfo{person}{Michael~S Brown}.}
  \bibinfo{year}{2014}\natexlab{}.
\newblock \showarticletitle{Illuminant estimation for color constancy: why
  spatial-domain methods work and the role of the color distribution}.
\newblock \bibinfo{journal}{\emph{JOSA A}} \bibinfo{volume}{31},
  \bibinfo{number}{5} (\bibinfo{year}{2014}), \bibinfo{pages}{1049--1058}.
\newblock


\bibitem[\protect\citeauthoryear{Cheng, Price, Cohen, and Brown}{Cheng
  et~al\mbox{.}}{2015}]%
        {cheng2015effective}
\bibfield{author}{\bibinfo{person}{Dongliang Cheng}, \bibinfo{person}{Brian
  Price}, \bibinfo{person}{Scott Cohen}, {and} \bibinfo{person}{Michael~S
  Brown}.} \bibinfo{year}{2015}\natexlab{}.
\newblock \showarticletitle{Effective learning-based illuminant estimation
  using simple features}. In \bibinfo{booktitle}{\emph{Proceedings of the IEEE
  Conference on Computer Vision and Pattern Recognition}}.
  \bibinfo{pages}{1000--1008}.
\newblock


\bibitem[\protect\citeauthoryear{Dong, Loy, He, and Tang}{Dong
  et~al\mbox{.}}{2016}]%
        {dong2016image}
\bibfield{author}{\bibinfo{person}{Chao Dong}, \bibinfo{person}{Chen~Change
  Loy}, \bibinfo{person}{Kaiming He}, {and} \bibinfo{person}{Xiaoou Tang}.}
  \bibinfo{year}{2016}\natexlab{}.
\newblock \showarticletitle{Image super-resolution using deep convolutional
  networks}.
\newblock \bibinfo{journal}{\emph{IEEE transactions on pattern analysis and
  machine intelligence}} \bibinfo{volume}{38}, \bibinfo{number}{2}
  (\bibinfo{year}{2016}), \bibinfo{pages}{295--307}.
\newblock


\bibitem[\protect\citeauthoryear{Fattal, Lischinski, and Werman}{Fattal
  et~al\mbox{.}}{2002}]%
        {fattal2002gradient}
\bibfield{author}{\bibinfo{person}{Raanan Fattal}, \bibinfo{person}{Dani
  Lischinski}, {and} \bibinfo{person}{Michael Werman}.}
  \bibinfo{year}{2002}\natexlab{}.
\newblock \showarticletitle{Gradient domain high dynamic range compression}. In
  \bibinfo{booktitle}{\emph{ACM Transactions on Graphics (TOG)}},
  Vol.~\bibinfo{volume}{21}. ACM, \bibinfo{pages}{249--256}.
\newblock


\bibitem[\protect\citeauthoryear{Finlayson, Hordley, and Hubel}{Finlayson
  et~al\mbox{.}}{2001}]%
        {finlayson2001color}
\bibfield{author}{\bibinfo{person}{Graham~D. Finlayson},
  \bibinfo{person}{Steven~D. Hordley}, {and} \bibinfo{person}{Paul~M. Hubel}.}
  \bibinfo{year}{2001}\natexlab{}.
\newblock \showarticletitle{Color by correlation: A simple, unifying framework
  for color constancy}.
\newblock \bibinfo{journal}{\emph{IEEE Transactions on Pattern Analysis and
  Machine Intelligence}} \bibinfo{volume}{23}, \bibinfo{number}{11}
  (\bibinfo{year}{2001}), \bibinfo{pages}{1209--1221}.
\newblock


\bibitem[\protect\citeauthoryear{Finlayson and Trezzi}{Finlayson and
  Trezzi}{2004}]%
        {finlayson2004shades}
\bibfield{author}{\bibinfo{person}{Graham~D Finlayson} {and}
  \bibinfo{person}{Elisabetta Trezzi}.} \bibinfo{year}{2004}\natexlab{}.
\newblock \showarticletitle{Shades of gray and colour constancy}. In
  \bibinfo{booktitle}{\emph{Color and Imaging Conference}},
  Vol.~\bibinfo{volume}{2004}. Society for Imaging Science and Technology,
  \bibinfo{pages}{37--41}.
\newblock


\bibitem[\protect\citeauthoryear{Gehler, Rother, Blake, Minka, and
  Sharp}{Gehler et~al\mbox{.}}{2008}]%
        {gehler2008bayesian}
\bibfield{author}{\bibinfo{person}{Peter~Vincent Gehler},
  \bibinfo{person}{Carsten Rother}, \bibinfo{person}{Andrew Blake},
  \bibinfo{person}{Tom Minka}, {and} \bibinfo{person}{Toby Sharp}.}
  \bibinfo{year}{2008}\natexlab{}.
\newblock \showarticletitle{Bayesian color constancy revisited}. In
  \bibinfo{booktitle}{\emph{Computer Vision and Pattern Recognition, 2008. CVPR
  2008. IEEE Conference on}}. IEEE, \bibinfo{pages}{1--8}.
\newblock


\bibitem[\protect\citeauthoryear{Gijsenij and Gevers}{Gijsenij and
  Gevers}{2011}]%
        {gijsenij2011color}
\bibfield{author}{\bibinfo{person}{Arjan Gijsenij} {and} \bibinfo{person}{Theo
  Gevers}.} \bibinfo{year}{2011}\natexlab{}.
\newblock \showarticletitle{Color constancy using natural image statistics and
  scene semantics}.
\newblock \bibinfo{journal}{\emph{IEEE Transactions on Pattern Analysis and
  Machine Intelligence}} \bibinfo{volume}{33}, \bibinfo{number}{4}
  (\bibinfo{year}{2011}), \bibinfo{pages}{687--698}.
\newblock


\bibitem[\protect\citeauthoryear{Guo, Li, and Ling}{Guo et~al\mbox{.}}{2017}]%
        {guo2017lime}
\bibfield{author}{\bibinfo{person}{Xiaojie Guo}, \bibinfo{person}{Yu Li}, {and}
  \bibinfo{person}{Haibin Ling}.} \bibinfo{year}{2017}\natexlab{}.
\newblock \showarticletitle{LIME: Low-Light Image Enhancement via Illumination
  Map Estimation}.
\newblock \bibinfo{journal}{\emph{IEEE Transactions on Image Processing}}
  \bibinfo{volume}{26}, \bibinfo{number}{2} (\bibinfo{year}{2017}),
  \bibinfo{pages}{982--993}.
\newblock


\bibitem[\protect\citeauthoryear{He, Sun, and Tang}{He et~al\mbox{.}}{2009}]%
        {CVPR2009_He}
\bibfield{author}{\bibinfo{person}{Kaiming He}, \bibinfo{person}{Jian Sun},
  {and} \bibinfo{person}{Xiaoou Tang}.} \bibinfo{year}{2009}\natexlab{}.
\newblock \showarticletitle{Single image haze removal using dark channel
  prior}. In \bibinfo{booktitle}{\emph{Proceedings of the IEEE conference on
  computer vision and pattern recognition}}.
\newblock


\bibitem[\protect\citeauthoryear{He, Sun, and Tang}{He et~al\mbox{.}}{2011}]%
        {he2011single}
\bibfield{author}{\bibinfo{person}{Kaiming He}, \bibinfo{person}{Jian Sun},
  {and} \bibinfo{person}{Xiaoou Tang}.} \bibinfo{year}{2011}\natexlab{}.
\newblock \showarticletitle{Single image haze removal using dark channel
  prior}.
\newblock \bibinfo{journal}{\emph{IEEE transactions on pattern analysis and
  machine intelligence}} \bibinfo{volume}{33}, \bibinfo{number}{12}
  (\bibinfo{year}{2011}), \bibinfo{pages}{2341--2353}.
\newblock


\bibitem[\protect\citeauthoryear{He, Zhang, Ren, and Sun}{He
  et~al\mbox{.}}{2016}]%
        {he2016deep}
\bibfield{author}{\bibinfo{person}{Kaiming He}, \bibinfo{person}{Xiangyu
  Zhang}, \bibinfo{person}{Shaoqing Ren}, {and} \bibinfo{person}{Jian Sun}.}
  \bibinfo{year}{2016}\natexlab{}.
\newblock \showarticletitle{Deep residual learning for image recognition}. In
  \bibinfo{booktitle}{\emph{Proceedings of the IEEE conference on computer
  vision and pattern recognition}}. \bibinfo{pages}{770--778}.
\newblock


\bibitem[\protect\citeauthoryear{Hirschmuller and Scharstein}{Hirschmuller and
  Scharstein}{2007}]%
        {hirschmuller2007evaluation}
\bibfield{author}{\bibinfo{person}{Heiko Hirschmuller} {and}
  \bibinfo{person}{Daniel Scharstein}.} \bibinfo{year}{2007}\natexlab{}.
\newblock \showarticletitle{Evaluation of cost functions for stereo matching}.
  In \bibinfo{booktitle}{\emph{Computer Vision and Pattern Recognition, 2007.
  CVPR'07. IEEE Conference on}}. IEEE, \bibinfo{pages}{1--8}.
\newblock


\bibitem[\protect\citeauthoryear{Howard, Zhu, Chen, Kalenichenko, Wang, Weyand,
  Andreetto, and Adam}{Howard et~al\mbox{.}}{2017}]%
        {howard2017mobilenets}
\bibfield{author}{\bibinfo{person}{Andrew~G Howard}, \bibinfo{person}{Menglong
  Zhu}, \bibinfo{person}{Bo Chen}, \bibinfo{person}{Dmitry Kalenichenko},
  \bibinfo{person}{Weijun Wang}, \bibinfo{person}{Tobias Weyand},
  \bibinfo{person}{Marco Andreetto}, {and} \bibinfo{person}{Hartwig Adam}.}
  \bibinfo{year}{2017}\natexlab{}.
\newblock \showarticletitle{Mobilenets: Efficient convolutional neural networks
  for mobile vision applications}.
\newblock \bibinfo{journal}{\emph{arXiv preprint arXiv:1704.04861}}
  (\bibinfo{year}{2017}).
\newblock


\bibitem[\protect\citeauthoryear{Hu, Wang, and Lin}{Hu et~al\mbox{.}}{2017}]%
        {hu2017fc}
\bibfield{author}{\bibinfo{person}{Yuanming Hu}, \bibinfo{person}{Baoyuan
  Wang}, {and} \bibinfo{person}{Stephen Lin}.} \bibinfo{year}{2017}\natexlab{}.
\newblock \showarticletitle{FC4: Fully Convolutional Color Constancy with
  Confidence-weighted Pooling}. In \bibinfo{booktitle}{\emph{Proceedings of the
  IEEE Conference on Computer Vision and Pattern Recognition}}.
  \bibinfo{pages}{4085--4094}.
\newblock


\bibitem[\protect\citeauthoryear{Jia, Shelhamer, Donahue, Karayev, Long,
  Girshick, Guadarrama, and Darrell}{Jia et~al\mbox{.}}{2014}]%
        {jia2014caffe}
\bibfield{author}{\bibinfo{person}{Yangqing Jia}, \bibinfo{person}{Evan
  Shelhamer}, \bibinfo{person}{Jeff Donahue}, \bibinfo{person}{Sergey Karayev},
  \bibinfo{person}{Jonathan Long}, \bibinfo{person}{Ross Girshick},
  \bibinfo{person}{Sergio Guadarrama}, {and} \bibinfo{person}{Trevor Darrell}.}
  \bibinfo{year}{2014}\natexlab{}.
\newblock \showarticletitle{Caffe: Convolutional architecture for fast feature
  embedding}. In \bibinfo{booktitle}{\emph{Proceedings of the 22nd ACM
  international conference on Multimedia}}. ACM, \bibinfo{pages}{675--678}.
\newblock


\bibitem[\protect\citeauthoryear{Joze, Drew, Finlayson, and Rey}{Joze
  et~al\mbox{.}}{2012}]%
        {joze2012role}
\bibfield{author}{\bibinfo{person}{Hamid Reza~Vaezi Joze},
  \bibinfo{person}{Mark~S Drew}, \bibinfo{person}{Graham~D Finlayson}, {and}
  \bibinfo{person}{Perla Aurora~Troncoso Rey}.}
  \bibinfo{year}{2012}\natexlab{}.
\newblock \showarticletitle{The role of bright pixels in illumination
  estimation}. In \bibinfo{booktitle}{\emph{Color and Imaging Conference}},
  Vol.~\bibinfo{volume}{2012}. Society for Imaging Science and Technology,
  \bibinfo{pages}{41--46}.
\newblock


\bibitem[\protect\citeauthoryear{Karklin and Lewicki}{Karklin and
  Lewicki}{2005}]%
        {karklin2005hierarchical}
\bibfield{author}{\bibinfo{person}{Yan Karklin} {and}
  \bibinfo{person}{Michael~S Lewicki}.} \bibinfo{year}{2005}\natexlab{}.
\newblock \showarticletitle{A hierarchical Bayesian model for learning
  nonlinear statistical regularities in nonstationary natural signals}.
\newblock \bibinfo{journal}{\emph{Neural computation}} \bibinfo{volume}{17},
  \bibinfo{number}{2} (\bibinfo{year}{2005}), \bibinfo{pages}{397--423}.
\newblock


\bibitem[\protect\citeauthoryear{Li, Peng, Wang, Xu, and Feng}{Li
  et~al\mbox{.}}{2017}]%
        {li2017all}
\bibfield{author}{\bibinfo{person}{Boyi Li}, \bibinfo{person}{Xiulian Peng},
  \bibinfo{person}{Zhangyang Wang}, \bibinfo{person}{Jizheng Xu}, {and}
  \bibinfo{person}{Dan Feng}.} \bibinfo{year}{2017}\natexlab{}.
\newblock \showarticletitle{Aod-net: All-in-one dehazing network}. In
  \bibinfo{booktitle}{\emph{Proceedings of the IEEE International Conference on
  Computer Vision}}, Vol.~\bibinfo{volume}{1}. \bibinfo{pages}{7}.
\newblock


\bibitem[\protect\citeauthoryear{Peng and Cosman}{Peng and Cosman}{2017}]%
        {peng2017underwater}
\bibfield{author}{\bibinfo{person}{Yan-Tsung Peng} {and}
  \bibinfo{person}{Pamela~C Cosman}.} \bibinfo{year}{2017}\natexlab{}.
\newblock \showarticletitle{Underwater Image Restoration Based on Image
  Blurriness and Light Absorption}.
\newblock \bibinfo{journal}{\emph{IEEE Transactions on Image Processing}}
  \bibinfo{volume}{26}, \bibinfo{number}{4} (\bibinfo{year}{2017}),
  \bibinfo{pages}{1579--1594}.
\newblock


\bibitem[\protect\citeauthoryear{Ren, Liu, Zhang, Pan, Cao, and Yang}{Ren
  et~al\mbox{.}}{2016}]%
        {ren2016single}
\bibfield{author}{\bibinfo{person}{Wenqi Ren}, \bibinfo{person}{Si Liu},
  \bibinfo{person}{Hua Zhang}, \bibinfo{person}{Jinshan Pan},
  \bibinfo{person}{Xiaochun Cao}, {and} \bibinfo{person}{Ming-Hsuan Yang}.}
  \bibinfo{year}{2016}\natexlab{}.
\newblock \showarticletitle{Single image dehazing via multi-scale convolutional
  neural networks}. In \bibinfo{booktitle}{\emph{European Conference on
  Computer Vision}}. Springer, \bibinfo{pages}{154--169}.
\newblock


\bibitem[\protect\citeauthoryear{Shi}{Shi}{2000}]%
        {shi2000re}
\bibfield{author}{\bibinfo{person}{Lilong Shi}.}
  \bibinfo{year}{2000}\natexlab{}.
\newblock \showarticletitle{Re-processed version of the gehler color constancy
  dataset of 568 images}.
\newblock \bibinfo{journal}{\emph{http://www.cs.sfu.ca/\%7Ecolour/data/,}}
  (\bibinfo{year}{2000}).
\newblock


\bibitem[\protect\citeauthoryear{Shi, Loy, and Tang}{Shi et~al\mbox{.}}{2016}]%
        {shi2016deep}
\bibfield{author}{\bibinfo{person}{Wu Shi}, \bibinfo{person}{Chen~Change Loy},
  {and} \bibinfo{person}{Xiaoou Tang}.} \bibinfo{year}{2016}\natexlab{}.
\newblock \showarticletitle{Deep specialized network for illuminant
  estimation}. In \bibinfo{booktitle}{\emph{European Conference on Computer
  Vision}}. Springer, \bibinfo{pages}{371--387}.
\newblock


\bibitem[\protect\citeauthoryear{Szegedy, Liu, Jia, Sermanet, Reed, Anguelov,
  Erhan, Vanhoucke, and Rabinovich}{Szegedy et~al\mbox{.}}{2015}]%
        {szegedy2015going}
\bibfield{author}{\bibinfo{person}{Christian Szegedy}, \bibinfo{person}{Wei
  Liu}, \bibinfo{person}{Yangqing Jia}, \bibinfo{person}{Pierre Sermanet},
  \bibinfo{person}{Scott Reed}, \bibinfo{person}{Dragomir Anguelov},
  \bibinfo{person}{Dumitru Erhan}, \bibinfo{person}{Vincent Vanhoucke}, {and}
  \bibinfo{person}{Andrew Rabinovich}.} \bibinfo{year}{2015}\natexlab{}.
\newblock \showarticletitle{Going deeper with convolutions}. In
  \bibinfo{booktitle}{\emph{Proceedings of the IEEE conference on computer
  vision and pattern recognition}}. \bibinfo{pages}{1--9}.
\newblock


\bibitem[\protect\citeauthoryear{Tang, Yang, and Wang}{Tang
  et~al\mbox{.}}{2014}]%
        {tang2014investigating}
\bibfield{author}{\bibinfo{person}{Ketan Tang}, \bibinfo{person}{Jianchao
  Yang}, {and} \bibinfo{person}{Jue Wang}.} \bibinfo{year}{2014}\natexlab{}.
\newblock \showarticletitle{Investigating haze-relevant features in a learning
  framework for image dehazing}. In \bibinfo{booktitle}{\emph{Proceedings of
  the IEEE Conference on Computer Vision and Pattern Recognition}}.
  \bibinfo{pages}{2995--3000}.
\newblock


\bibitem[\protect\citeauthoryear{Van De~Weijer, Gevers, and Gijsenij}{Van
  De~Weijer et~al\mbox{.}}{2007}]%
        {van2007edge}
\bibfield{author}{\bibinfo{person}{Joost Van De~Weijer}, \bibinfo{person}{Theo
  Gevers}, {and} \bibinfo{person}{Arjan Gijsenij}.}
  \bibinfo{year}{2007}\natexlab{}.
\newblock \showarticletitle{Edge-based color constancy}.
\newblock \bibinfo{journal}{\emph{IEEE Transactions on image processing}}
  \bibinfo{volume}{16}, \bibinfo{number}{9} (\bibinfo{year}{2007}),
  \bibinfo{pages}{2207--2214}.
\newblock


\bibitem[\protect\citeauthoryear{Xie, Xu, and Chen}{Xie et~al\mbox{.}}{2012}]%
        {xie2012image}
\bibfield{author}{\bibinfo{person}{Junyuan Xie}, \bibinfo{person}{Linli Xu},
  {and} \bibinfo{person}{Enhong Chen}.} \bibinfo{year}{2012}\natexlab{}.
\newblock \showarticletitle{Image denoising and inpainting with deep neural
  networks}. In \bibinfo{booktitle}{\emph{Advances in Neural Information
  Processing Systems}}. \bibinfo{pages}{341--349}.
\newblock


\bibitem[\protect\citeauthoryear{Zhang, Cao, Fang, Kang, and Chen}{Zhang
  et~al\mbox{.}}{2017a}]%
        {zhang2017fast}
\bibfield{author}{\bibinfo{person}{Jing Zhang}, \bibinfo{person}{Yang Cao},
  \bibinfo{person}{Shuai Fang}, \bibinfo{person}{Yu Kang}, {and}
  \bibinfo{person}{Chang~Wen Chen}.} \bibinfo{year}{2017}\natexlab{a}.
\newblock \showarticletitle{Fast Haze Removal for Nighttime Image Using Maximum
  Reflectance Prior}. In \bibinfo{booktitle}{\emph{Proceedings of the IEEE
  Conference on Computer Vision and Pattern Recognition}}.
  \bibinfo{pages}{7418--7426}.
\newblock


\bibitem[\protect\citeauthoryear{Zhang, Zuo, Chen, Meng, and Zhang}{Zhang
  et~al\mbox{.}}{2017c}]%
        {zhang2017beyond}
\bibfield{author}{\bibinfo{person}{Kai Zhang}, \bibinfo{person}{Wangmeng Zuo},
  \bibinfo{person}{Yunjin Chen}, \bibinfo{person}{Deyu Meng}, {and}
  \bibinfo{person}{Lei Zhang}.} \bibinfo{year}{2017}\natexlab{c}.
\newblock \showarticletitle{Beyond a gaussian denoiser: Residual learning of
  deep cnn for image denoising}.
\newblock \bibinfo{journal}{\emph{IEEE Transactions on Image Processing}}
  (\bibinfo{year}{2017}).
\newblock


\bibitem[\protect\citeauthoryear{Zhang, Zhou, Lin, and Sun}{Zhang
  et~al\mbox{.}}{2017b}]%
        {zhang2017shufflenet}
\bibfield{author}{\bibinfo{person}{Xiangyu Zhang}, \bibinfo{person}{Xinyu
  Zhou}, \bibinfo{person}{Mengxiao Lin}, {and} \bibinfo{person}{Jian Sun}.}
  \bibinfo{year}{2017}\natexlab{b}.
\newblock \showarticletitle{Shufflenet: An extremely efficient convolutional
  neural network for mobile devices}.
\newblock \bibinfo{journal}{\emph{arXiv preprint arXiv:1707.01083}}
  (\bibinfo{year}{2017}).
\newblock


\bibitem[\protect\citeauthoryear{Zhu, Mai, and Shao}{Zhu et~al\mbox{.}}{2014}]%
        {zhu2014single}
\bibfield{author}{\bibinfo{person}{Qingsong Zhu}, \bibinfo{person}{Jiaming
  Mai}, {and} \bibinfo{person}{Ling Shao}.} \bibinfo{year}{2014}\natexlab{}.
\newblock \showarticletitle{Single Image Dehazing Using Color Attenuation
  Prior.}. In \bibinfo{booktitle}{\emph{BMVC}}.
\newblock


\end{thebibliography}

\end{document}